\documentclass[a4]{article}
\usepackage{cite}
\usepackage{amsmath}
\usepackage{amssymb}
\usepackage{bbm}
\usepackage[linesnumbered,boxed]{algorithm2e}
\SetAlCapSkip{\baselineskip}

\SetCommentSty{SmallCommentFont}
\usepackage{textcomp}
\usepackage{graphicx,float}
\usepackage{ifpdf}
\usepackage{ifthen}
\ifpdf
\usepackage[bookmarksnumbered]{hyperref}
\else
\usepackage[bookmarks=false]{hyperref}
\fi
\SetKwInput{KwInput}{Input}
\SetKwInput{KwOutput}{Output}
\SetKwInput{KwInputOutput}{Input/Output}
\newcommand{\Mr}[1]{\mathrm{#1}}%
\newcommand{\Ms}[1]{\mathsf{#1}}%
\newcommand{\Ts}[1]{\textsf{#1}}%
\newcommand{\Tb}[1]{\textbf{#1}}%
\newcommand{\Em}[1]{\emph{#1}}

\newcommand{\Ba}{\begin{align}}%
\newcommand{\Bc}{\begin{center}}%
\newcommand{\Ec}{\end{center}}%
\newcommand{\Be}{\begin{enumerate}}%
\newcommand{\Ee}{\end{enumerate}}%
\newcommand{\Bi}{\begin{itemize}}%
\newcommand{\Ei}{\end{itemize}}%
\newcommand{\MbB}{\ensuremath{\mathbb{B}}}

\newcommand{\MbN}{\ensuremath{\mathbb{N}}}

\newcommand{\MbQ}{\ensuremath{\mathbb{Q}}}
\newcommand{\MbR}{\ensuremath{\mathbb{R}}}

\newcommand{\McO}{\ensuremath{\mathcal{O}}}
\newcommand{\NumBins}{\mathsf{nb}}

\begin{document}
\newcommand{\nographics}{0}
\newcommand{\usegraphics}[5]
%
%
%
%
{\ifthenelse{\nographics = 1}
{\begin{center}\fbox{\fbox{\parbox[t]{0.25\textwidth}
{\textsf{\huge X}\begin{center}
\ifpdf{#5.pdf}\else{#5.eps}\fi
\end{center}}}}
\end{center}}
{\ifpdf
{\rotatebox{#2}{\resizebox{#3}{#4}{\includegraphics{#5.pdf}}}}
\else{\rotatebox{#1}{\resizebox{#3}{#4}{\includegraphics{#5.eps}}}}
\fi}}
\title{Defect Detection on Semiconductor Wafers by Distribution Analysis}
\date{\today}
\author{Thomas Olschewski\footnote{thomas.olschewski@tu-dresden.de},
Technische Universit\"at Dresden}
\begin{titlepage}
\maketitle
\begin{abstract}
  A method for object classification that is based on distribution analysis
  is proposed.  In addition, a method for finding relevant features and the
  unification of this algorithm with another classification algorithm is
  proposed.  The presented classification algorithm has been applied
  successfully to real-world measurement data from wafer fabrication of close
  to hundred thousand chips of several product types.  The presented algorithm
  prefers finding the best rater in a low-dimensional search space over
  finding a good rater in a high-dimensional search space.  Our approach is
  interesting in that it is fast (quasi-linear) and reached good to excellent
  prediction or detection quality for real-world wafer data.
\end{abstract}
\end{titlepage}

\tableofcontents
\section{Acknowledgements}
\label{sec:Acknowledgements}
I would like to thank Zolt\'{a}n Sasv\'{a}ri for carefully reading the
manuscript and for providing the environment which made this research
possible.
\\~\\
Research leading to these results has received funding from the iRel40
project.  iRel40 is a European co-funded innovation project that has been
granted by the \Tb{ECSEL} Joint Undertaking (JU) under grant agreement
\Tb{n$^\mathbf{o}$ 876659}. The funding of the project comes from the Horizon
2020 research programme and participating countries. National funding is
provided by Germany, including the Free States of Saxony and Thuringia,
Austria, Belgium, Finland, France, Italy, the Netherlands, Slovakia, Spain,
Sweden, and Turkey.
\\~\\
\Tb{Disclaimer:} The document reflects only the author's view and the JU is
not responsible for any use that may be made of the information it contains.
%
%
\begin{figure}[H]
\usegraphics{270}{0}{0.4\textwidth}{0.1\textwidth}{logo-ECSEL-1}
\qquad
\usegraphics{270}{0}{0.16\textwidth}{0.1125\textwidth}{flag-yellow-cropped}
\qquad
\usegraphics{270}{0}{0.15\textwidth}{0.1\textwidth}{logo-irel40}
\qquad
\usegraphics{270}{0}{0.2\textwidth}{0.1\textwidth}{FS-LM-poweredby}
\qquad
\usegraphics{270}{0}{0.3\textwidth}{0.2\textwidth}{BMBF-CMYK-Gef-L-e}
\end{figure}
%
\section{Introduction}
\label{sec:Introduction}
In this paper we propose a fast stochastic method for object classification
which can be applied to detecting and predicting defective semiconductor
devices by analyzing high-dimensional measurement data from wafer fabrication.
The most important goal is predicting early in the production chain which
semiconductor chips of a set of wafers will turn out to be defective later.
\par
Our approach is interesting in that it led to good to excellent results in
frontend classification tasks with real-world wafer data and is based on
distributional analysis of the input data with quasilinear time complexity
displaying computation times of some minutes for typical classification tasks
of ours on normal PC hardware.
\par
The function principles we are using in the classification algorithm presented
in this paper are deriving certain distributional properties from the scaled
features of two sets of positive and negative samples in order to select a
subset of features with specific properties as candidate indicator features.
These latter features are then used for attempting to classify objects
hitherto unknown to the algorithm. 
\par
We will describe example applications of how this algorithm is used for
detecting frontend defects using only tiny sample sets.  Another application
we will present here is trying to predict backend defects using only a very
small number of frontend measurements.
\par
Part of the tasks we are tackling here have also been attacked by the
algorithm described in \cite{Olschewski:2021} so we can compare the quality of
detection or prediction by these two algorithms which are based on completely
different principles.  Similar as with the latter algorithm and unlike, for
example, typical neural-net classification methods\cite{Rumelhart:1986} with
back-propagation, we try to avoid optimizations over high-dimensional search
ranges and prefer finding some best rater in a limited set of raters over
trying to find a good rater in a giant search range.
\par
The algorithms presented in this paper have quasi-linear time complexity.
They have been implemented in Python and have been used successfully for
classifying close to hundred thousand semiconductor chips of different product
types, where classifying 10000 chips took some minutes on normal PC hardware.
We expect that re-implementation in C/C++ and parallelization would bring a
considerable speedup.
\par
The main application of the classification algorithm described in this paper
is improving quality control in wafer fabrication.  In an ongoing series of
research projects working towards this goal, the following
properties---already mentioned in \cite{Olschewski:2021}---have turned out to
be important for some algorithm to be economically useful:
\begin{enumerate}
\item Not dependent on Gaussian distribution of some or all features.
\item High TP/FP quotients: as few as possible good chips should be scrapped
  for sorting out defective chips.
\item Efficiency: necessity of classification in fabrication real time.
\item Ability of coping with large amount of data, for example
  $\approx 30-100$ MiB per data lot.
\item Means for reducing the number of tests: measurements are costly and time
  consuming to different extents.
\end{enumerate}
Algorithm \ref{alg:Classification} and its auxiliary functions explained in
section \ref{sec:TheAlgorithms} have been designed to meet these goals.
\par
In some sense, this paper can be seen as a follow-up to
\cite{Olschewski:2021}. Whereas in the latter paper we reported on frontend
detection and prediction tasks only, in this paper we also consider backend
predictions based on frontend measurements, using a classification method
based on completely different principles.
\subsection{Overview}
\label{ssec:Overview}
The rest of this paper is organized as follows.  In section
\ref{sec:TheAlgorithms} we will define notations and describe our algorithms.
In section \ref{sec:ExperimentalSetup} we describe the experimental setup, the
tasks to be solved and the data material.  In section \ref{sec:Results} we
will list and analyse a number of results we obtained from a larger number of
runs attempting to classify semiconductor chips with respect to the overall
defect state of one of several possible measurement steps as the main
application, and results from classifying iris data.  We will also propose and
apply a feature relevance indicator in section \ref{sec:Results}.  Section
\ref{sec:AlgorithmicComplexity} will be about the worst-case complexity of the
algorithms described in section \ref{sec:TheAlgorithms}.  In section
\ref{sec:Uniting} we will propose an algorithm which unites the classification
Algorithm \ref{alg:Classification} of this paper with the classification
algorithm described in \cite{Olschewski:2021} by implanting some main
ingredients of the latter algorithm into the former.  Section
\ref{sec:Conclusion} will close this paper by concluding thoughts.

\section{The Algorithms}
\label{sec:TheAlgorithms}
\subsection{Definitions}
\label{ssec:Definitions}
In what follows, we will use the abbreviations $\MbB=\{0,1\}$,
$\MbQ^{\ast}=\MbQ\cup\{-\infty,\infty\}$.
\par
Let $X\subseteq\MbQ^{m\times n}$ be a data matrix describing $m$ objects by
$n$ numbers each.
\subsubsection{Scaling}
\label{sssec:Scaling}
To begin with, the first step of our classification algorithm is column-wise
scaling the input data matrix $X$.  Scale($X$) refers to the following
function.
\\
\begin{algorithm}[H]
  \label{alg:Scale}
  \KwInputOutput{$X\in\MbQ^{m\times n}$}
  \caption{Scale($X$)}
  \For{$j=1\ldots n$}
  {
    compute $\mu_{j},\sigma_{j}$\\
    \tcp{mean and standard deviation of the $j$-th column
       of $X$}
  }
  \For{$i=1\ldots m$}
  {
    \For{$j=1\ldots n$}
    {
      $x_{i,j} = \begin{cases}
               \frac{x_{i,j}-\mu_{j}}{\sigma_{j}},
                   & \sigma_{j}\neq0\\
               0, & \sigma_{j}=0
               \end{cases}$
    }
  }
\end{algorithm}
~\\
Every column of $X$ has mean 0 and standard deviation 1 after scaling.  After
scaling $X$, the algorithms do not operate on the original rows
$x_{i,1},\ldots,x_{i,n})$ but on the corresponding numbers that quantify how
far (by how many standard deviations $\sigma_{j}$) $x_{i,j}$ is away from the
mean $\mu_{j}$ of its column $j$.
\par
\Tb{Note.} We always assume $\Mr{Scale}(X)\in\MbQ^{m\times n}$.  This is the
case if $\Mr{Scale}(X)$ is computed numerically using floating point
arithmetics.  With symbolic computation, the element type of $\Mr{Scale}(X)$
would be an extension field $\MbQ(\sigma_{1},\ldots,\sigma_{n})$.
\subsubsection{Histogram Functions}
\label{sssec:HistogramFunctions}
We consider the input objects---represented by lines of the input data matrix
$X$---as samples of random processes the true distributions of which are
unknown to us.  Furthermore, we consider positive objects as results of a
random process generating positive objects and negative objects as results of
a possibly different random process generating negative objects.  For guessing
certain statistical properties of the true distributions, we need a means to
re-construct approximations of the shape of the true distributions.  For doing
this, we use histograms derived from the sample objects $T^{+},T^{-}$ which
will be described now.
\par
Let $v\in\MbB^{m}$ be a 0-1 column vector.  We call the
object belonging to line $i$ a \Tb{positive object} if $v_{i}=1$ and a
\Tb{negative object} if $v_{i}=0$.  Let
\[  I^{+}=\{i\in1,\ldots,m\mid v_{i}=1\}
\]
and
\[  I^{-}=\{i\in1,\ldots,m\mid v_{i}=0\}
\]
be the sets of line indices belonging to all positive resp. negative objects.
Let $T^{+}\subseteq I^{+}$ and $T^{-}\subseteq I^{-}$ be two subsets which
will serve as index sets of a set of \Tb{positive training objects} and a set
of \Tb{negative training objects}.
Let be
\[  I_{\neg T} = \{1,\ldots,m\}\setminus (T^{+}\cup T^{-})
\]
the index set of objects outside of the training sets.  This is the set of
indices of all objects to be classified.  We will skip training objects in the
classification loop for two reasons: firstly, in order to keep a clean
separation of what is known to the algorithm ($v(i)$ for $i\in I_{\neg T}$)
and what is to be predicted ($v(i)$ for $i\not\in T^{+}\cup T^{-}$). And
secondly, we consider classifying objects already known to the algorithm as a
different, simpler problem than classifying objects not yet known to the
algorithm.  Later, we will also need the index set of all training objects
\[ I_{T} = T^{+}\cup T^{-}.
\]
\par
Since what we actually will be doing is using these two sets of training
objects as samples for guessing certain statistical properties of the full
distributions of the input data, we can call the two training sets \Tb{positive
  samples} and \Tb{negative samples} as well.
\par
For convenience and if there is no danger of misinterpretation, we will not
always make a difference when referring to objects and referring to lines of
$X$.
\par
We assume $\NumBins\geqq3$ (``number of bins'') to be a constant which will be used
for setting the number of bins in computing all occurring histograms.

Let $a$ be an increasing sequence of numbers
\[
  a=(-\infty < a_{1} < a_{2} < \cdots < a_{\NumBins-1} < \infty)
     \in\{-\infty\}\times{\MbQ^{\ast}}^{\NumBins-1}\times\{\infty\}
\]
whereby we always set $a_{0}=-\infty$, $a_{\NumBins}=\infty$.
\par
Every such $a$ induces a partitioning
\[ \MbQ^{\ast} = \sqcup_{k=0}^{\NumBins-1} I_{i}
\]
into mostly half-open intervals\footnote{The only exception being the first
  interval $I_{0}=(-\infty,a_{1})$ which is open at both ends.} where
\[
  \begin{array}{lclcl}
    I_{0} &=& (a_{0},a_{1}) &=& (-\infty,a_{1}) \\
    I_{1} &=& [a_{1},a_{2}) \\
    \ldots && \ldots \\
    I_{i} &=& [a_{i},a_{i+1}) && (i=0,\ldots,\NumBins-1)\\
    \ldots \\
    I_{\NumBins-2} &=& [a_{\NumBins-2},a_{\NumBins-1}) \\
    I_{\NumBins-1} &=& [a_{\NumBins-1},a_{\NumBins}) &=& [a_{\NumBins-1},\infty)
  \end{array}
\]
The assumption $\NumBins\geqq3$ guarantees that there is at least one interval
$[a_{1},a_{2})$ with both limits being finite.
\par
\Tb{Definition.}\label{def:H} Let
$\mathrm{H}_{a,\NumBins}:\MbQ^{m}\to\MbQ^{\NumBins}$ be the
histogram function that, given a sequence of numbers $x=(x_{1},\ldots,x_{m})$
and a sequence of boundaries
$a=(-\infty < a_{1} < a_{2} < \cdots < a_{\NumBins-1} < \infty)$, assigns to
every bin index $k\in\{0,1,\ldots,\NumBins-1\}$ the relative frequency of this
bin, i.e. the number of elements lying in $I_{k}$ divided by $m$:
\[
    H_{a,\NumBins}(x)=(h_{0},\ldots,h_{\NumBins-1}) 
\]
where
\[
  h_{k}=\frac{1}{m}\cdot\big|\{ i\in\{1,\ldots,m\}\mid x_{i}\in I_{k} \}\big|
    \quad(k=0,1,\ldots,\NumBins-1)
\]
Obviously, $\sum_{k=0}^{\NumBins-1}h_{k}=1$.
Note that $a$ is required to be in strictly ascending order whereas $x$ may be
unordered.

Given $\NumBins$ and $x\in\MbQ^{m}$, we can specify boundaries
$a^{\ast}=(-\infty < a^{\ast}_{1} < a^{\ast}_{2} < \cdots <
a^{\ast}_{\NumBins-1} < \infty)$ in a way that the inner intervalls
$I^{\ast}_{1},\ldots,I^{\ast}_{\NumBins-2}$ have equal width
$\frac{\max(x)-\min(x)}{\NumBins-2}$.  We just set $a^{\ast}_{1}=\min(x)$ and
$a^{\ast}_{\NumBins-1}=\max(x)+\varepsilon$ for some $\varepsilon>0$.  Then
\[
    [\min(x),\max(x)+\varepsilon) = \sqcup_{k=1}^{\NumBins-2} I^{\ast}_{i}
\]
When using these boundaries $a^{\ast}$, the outmost intervals
$I_{0}=(-\infty,a^{\ast}_{1})$ and
$I_{\NumBins-1}=[a^{\ast}_{\NumBins-1},\infty)$ do not contain any element of
$x$ and thus $h_{0}=h_{\NumBins-1}=0$ in this case.  Every
$\NumBins$ setting induces a function which maps $x\in\MbQ^{m}$ to the
special boundaries vector $a^{\ast}$ defined above:
\begin{eqnarray*}
  a^{\ast}_{\NumBins}:\MbQ^{m} &\longrightarrow& {\MbQ^{\ast}}^{\NumBins+1} \\
  a^{\ast}_{\NumBins}(x) &=& (-\infty, a^{\ast}_{1}, a^{\ast}_{2}, \ldots,
    a^{\ast}_{\NumBins-1},\infty)
\end{eqnarray*}
\par
\Tb{Proposition.} Let $j\in\{1,\ldots,n\}$ be some column index. Let
$x_{1,j},\ldots,x_{m,j}$ be independent and identically distributed with
distribution $P^{(j)}$.  Let
$\mathcal{I}=\{i_{1},\ldots,i_{s}\}\subseteq\{1,\ldots,m\}$ some fixed subset
of indices and
$H_{a,\NumBins}^{\mathcal{I}}:\MbQ^{s}\longrightarrow\MbQ^{\NumBins}$,
$H_{a,\NumBins}^{\mathcal{I}}(x_{i_{1},j},\ldots,x_{i_{s},j})=
(h^{\mathcal{I}}_{0},\ldots,h^{\mathcal{I}}_{\NumBins-1})$ the histogram of
$(x_{i_{1},j},\ldots,x_{i_{s},j})$.  Then $h^{\mathcal{I}}_{k}$ is an unbiased
estimator of $P\big[x_{i,j}\in[a_{k},a_{k+1})\big]$ for
$k=0,\ldots,\NumBins-1$.\footnote{As an exception, the left-most interval
  $(a_{0},a_{1})=(-\infty,a_{1})$ is open on the left.}
\par
\Tb{Proof.}  Let $k\in\{0,\ldots,\NumBins-1\}$ be fixed.  Define
$X_{i_{1}},\ldots,X_{i_{s}}$ by
\[
  X_{i_{g}}=\begin{cases}
  1, & x_{i_{g}}\in[a_{k},a_{k+1}) \\
  0, & \text{else}
\end{cases}
\]
Then by linearity of the expectation $E$:
\begin{eqnarray*}
  E\left[h^{\mathcal{I}}_{k}\right]
  = E\left[\frac{X_{i_{1}}+\cdots+X_{i_{s}}}{s}\right] 
  = E\left[X_{i_{1}}\right]
  = P\big[x_{i,j}\in[a_{k},a_{k+1})\big],\quad\Mr{qed}.
\end{eqnarray*}
Thus by using $T^{+}\cup T^{-}$ for $\mathcal{I}$, we can take histograms
derived from the training set as unbiased estimators for the distribution of
every column $j$.
\subsubsection{Finding Candidate Indicator Columns}
\label{sssec:FindingCandidateIndicatorColumns}
\begin{algorithm}[H]
  \label{alg:FindCics}
  \KwInput{$\Mr{Scale}(X)\in\MbQ^{m\times n},T^{+}\subseteq I^{+},
    T^{-}\subseteq I^{-}, b^{+}\in\MbQ,b^{-}\in\MbQ,\NumBins\in\MbN_{\geqq3}$}
  \KwOutput{list of Cics $L=[(j_{s},a^{\ast}_{k_{s}},a^{\ast}_{k_{s}+1})
    \mid s=1,\ldots,S]$}
  \caption{FindCics($X,T^{+},T^{-},b^{+},b^{-},\NumBins$)}
  \For{$j=1\ldots n$}
  {
    set $x^{+}=(x_{i,j}\mid i\in T^{+})$ \\
    compute $a^{\ast}=a^{\ast}_{\NumBins}(x^{+})$ \\
    compute
      $h^{+}=H_{a^{\ast},\NumBins}(x^{+})\in\MbQ^{\NumBins}$ \\
    $h^{+}_{\max}=\max(h^{+}_{0},\ldots,h^{+}_{\NumBins-1})$ \\
    $k^{+}=\min\{k\in\{0,\ldots,\NumBins-1\}\mid h^{+}_{k}=h^{+}_{\max}\}$ \\
    set $x^{-}=(x_{i,j}\mid i\in T^{-})$ \\
    compute
      $h^{-}=H_{a^{\ast},\NumBins}(x^{-})\in\MbQ^{\NumBins}$ \\
    \If{$h^{+}_{\max}>b^{+}\wedge h^{-}_{k^{+}} < b^{-}$}
    { append $(j, a^{\ast}_{k^{+}}, a^{\ast}_{k^{+}+1})$ to $L$ }
  }
\end{algorithm}
~\\
$\Mr{FindCics}(X,T^{+},T^{-},b^{+},b^{-},\NumBins)$ takes as input
the scaled input data matrix $\Mr{Scale}(X)$, two indices sets of positive and
negative training objects, respectively, a lower bound $b^{+}$ and an upper
bound $b^{-}$, and the number of bins $\NumBins$ to be used in the histograms.
\par
\Tb{Definition.}\label{def:cic} A \Tb{candidate indicator column} (\Tb{Cic})
is any column $j$ of $\Mr{Scale}(X)$ which has the property that the
most-frequent bin $I_{k^{+}}$ of the column $j$ entries of all lines belonging
to positive samples is $>b^{+}$ whereas the same bin $I_{k^{+}}$ (i.e. with
the same boundaries) of the column $j$ entries of all lines belonging to
negative samples is $<b^{-}$.
\par
If there are multiple $k\in\{0,\ldots,\NumBins-1\}$ with
$h^{+}_{k} = h^{+}_{\max}$ then we take the one $k$ with the smallest index in
the above algorithm \ref{alg:FindCics}:
\[
  k^{+}=\min\{ k \mid h^{+}_{k} = h^{+}_{\max}\}.
\]
In $I_{k^{+}}=[a^{\ast}_{k^{+}},a^{\ast}_{k^{+}+1})$, the 2nd
and 3rd entry of the triplet $(j, a^{\ast}_{k^{+}}, a^{\ast}_{k^{+}+1})$ being
appended to $L$ are the lower and upper boundary of the most-frequent bin
$I_{k^{+}}$ of the histogram of positive objects $h^{+}$.
\par
Clearly,
$\Mr{FindCics}(\Mr{Scale}(X),T^{+},T^{-},b^{+},b^{-},\NumBins)$
returns the list of all Cics of $\Mr{Scale}(X)$.
\subsubsection{Computing Indicator Values $\Ms{S}_{\Ms{C}}(i)$}
\label{sssec:ComputingIndicatorValues}
\Tb{Definition.} Let
$\Ms{C}=\Mr{FindCics}(\Mr{Scale}(X),T^{+},T^{-},b^{+},b^{-},\NumBins)$ be the
output of Algorithm \ref{alg:FindCics}.  Then define
\begin{eqnarray*}
  \Ms{S}_{\Ms{C}}: I_{\neg T} &\longrightarrow& \MbN_{\geqq0}\\
  \Ms{S}_{\Ms{C}}(i) &=& |\{(j,a^{\ast}_{k^{+}},a^{\ast}_{k^{+}+1}) \in \Ms{C}
                     \mid x_{i,j}\in [a^{\ast}_{k^{+}},a^{\ast}_{k^{+}+1}) \}|
\end{eqnarray*}
In words, $\Ms{S}_{\Ms{C}}$ assigns to every index $i$ of a non-training
object $(x_{i,1},\ldots,x_{i,n})$ the number of those Cics $j$ for which the
$j$-th column entry of this object lies in the most-frequent bin of the
histogram of positive samples.
\\~\\
\begin{algorithm}[H]
  \label{alg:ComputingIndicatorValues}
  \KwInput{$\Ms{C}=\Mr{FindCics}(\Mr{Scale}(X),T^{+},T^{-},b^{+},b^{-},\NumBins)$}
  \KwOutput{$\Ms{S}_{\Ms{C}}(I_{\neg T})$}
  \caption{Computing indicator values $\Ms{S}_{\Ms{C}}(i)$}
  \For{$i\in I_{\neg T}$}
  {
    compute $\Ms{S}_{\Ms{C}}(i) = |\{(j,a^{\ast}_{k^{+}},a^{\ast}_{k^{+}+1}) \in \Ms{C}
                     \mid x_{i,j}\in [a^{\ast}_{k^{+}},a^{\ast}_{k^{+}+1}) \}|$
  }
\end{algorithm}
\subsubsection{Computing Predictions $F(i,c)$}
\label{sssec:ComputingPredictions}
After having computed an indicator value $\Ms{S}_{\Ms{C}}(i)$ for every object
$i$ outside the training sets we need to obtain binary predictions, 1 for
predicting object $i$ to be positive, 0 for negative:
\\~\\
\begin{algorithm}[H]
  \label{alg:ComputingPredictions}
  \KwInput{$\Ms{S}_{\Ms{C}}\big(I_{\neg T}\big)$
    as computed by Algorithm \ref{alg:ComputingIndicatorValues}, $c\geqq0$}
  \KwOutput{$\Ms{F}(c)\in\MbB^{|I_{\neg T}|}$}
  \caption{Computing predictions}
  \For{$i \in \{1,\ldots,m\}\setminus (T^{+}\cup T^{-})$ {\rm ($=I_{\neg T}$)}}
  {
    $F(i, c) = \begin{cases}   1, & \Ms{S}_{\Ms{C}}(i)\geqq c \\
                                0, & \Ms{S}_{\Ms{C}}(i) < c
                \end{cases}$
  }
\end{algorithm}
~\\
Of the wide range of methods for specifying $C$ we desribe just two.
\subsubsection{Cutoff Selection}
\label{sssec:CutoffSelection}
Let be
\begin{eqnarray*}
  I_{\neg T}^{+} &=& I^{+} \cap I_{\neg T} \\
  I_{\neg T}^{-} &=& I^{-} \cap I_{\neg T} 
\end{eqnarray*}
the index sets of objects to be classified which are positive or negative,
respectively, in truth.  Their average indicator values are:
\begin{eqnarray*}
  Av^{1} &=& \frac{1}{|I_{\neg T}^{+}|}
             \cdot\sum_{i\in I_{\neg T}^{+}} \Ms{S}_{\Ms{C}}(i) \\
  Av^{0} &=& \frac{1}{|I_{\neg T}^{-}|}
             \cdot\sum_{i\in I_{\neg T}^{-}} \Ms{S}_{\Ms{C}}(i)
\end{eqnarray*}
In what follows, we assume\footnote{If this does not hold true in a specific
  application then this may be an indication of insufficient training sets:
  not characteristic enough, or with contradictive
  positive/negative bits $v(i)$.}
$Av^{1}-Av^{0}\geqq0$.
\\\mbox{}\\
\Tb{Na\"ive cutoff selection}.
\\\mbox{}\\
\[
    c = \frac{Av^{1}-Av^{0}}{2}
\]
This better-than-nothing selection can be computed very fast but may make
predictions $F(i,c)$ far from what can be achieved with a more sophisticated
cutoff.
\\\mbox{}\\
\Tb{Optimizing cutoff for some statistical quantity $Q(v,w)$}.
\\\mbox{}\\
Let $\left(i^{1},\ldots,i^{|I_{\neg T}|}\right)$ be an enumeration of
$I_{\neg T}$ in some fixed order and
\[
    \Ms{v} = \big(v(i^{1}),\ldots,v(i^{|I_{\neg T}|})\big)
\]
the 0-1 vector of their true positive/negative states.  Let be
\[
    \Ms{F}(c) = \big(F(i^{1},c),\ldots,F(i^{|I_{\neg T}|},c)\big)
\]
the 0-1 vector of all $F(i,c)$ for $i\in I_{\neg T}$ using the same index
ordering as in $\Ms{v}$.  Furthermore, let $Q(v,w)$ be some statistical
quantity measuring similarity of two 0-1 vectors $v$ and $w$. This may be
accuracy\footnote{Accuracy is the amount of coincident bits of two $r$-bit
  vectors:
  \[ \mathrm{accu}(v,w)=\frac{1}{r}\cdot\sum_{i={1}}^{r}\big(v_{i}\cdot
    w_{i}+(1-v_{i})(1-w_{i})\big)
  \]
}
or---by considering $\Ms{F}(c)$ and
$\Ms{v}$ as two raters of the same feature
vector---Cohen's kappa\footnote{Cohen's Kappa\cite{Cohen:1960} is used for
  measuring agreement of two binary $r$-bit raters $v$ and $w$
  and defined as
  \[
    \mathrm{kappa}(v,w)=\frac{\mathrm{accu}(v,w)-p_{e}}{1-p_{e}}
  \]
  where
  \[
     p_{e}=\frac{1}{r^{2}}(n_{v_{i}=0}\cdot n_{w_{i}=0} + n_{v_{i}=1}\cdot
     n_{w_{i}=1}).
  \]
}, among other possibilities.
\par
Denote
\begin{eqnarray*}
  \Ms{S}_{\Ms{C}}^{\min}&=& \min\Ms{S}_{\Ms{C}}\left(I_{\neg T}\right)\\
  \Ms{S}_{\Ms{C}}^{\max}&=& \max\Ms{S}_{\Ms{C}}\left(I_{\neg T}\right)
\end{eqnarray*}
\begin{algorithm}[H]
  \label{alg:OptimizingCutoff}
  \KwInput{$\Ms{C}=\Mr{FindCics}(\Mr{Scale}(X),T^{+},T^{-},b^{+},b^{-},\NumBins)$}
  \KwOutput{$C_{opt}\in\MbQ,\Ms{F}(C_{opt})\in\MbB^{|I_{\neg T}|}$}
  \caption{Optimizing cutoff $c$}
  \For{$c=\Ms{S}_{\Ms{C}}^{\min},\Ms{S}_{\Ms{C}}^{\min}+1,
        \ldots,\Ms{S}_{\Ms{C}}^{\max}$}
  {
    compute $\Ms{F}(c)$ by Algorithm \ref{alg:ComputingPredictions}\\
    compute $Q\big(\Ms{F}(c),\Ms{v}\big)$
  }
  $Q_{opt} = \max\{Q\big(\Ms{F}(c),\Ms{v}\big)\mid c\in\mathcal{C}\}$\\
  $C_{opt}= \mathrm{argmax}(Q_{opt})$
\end{algorithm}
~
\par
Since the indicator values $\Ms{S}_{\Ms{C}}(i)$ are small integers which are
limited to the number of Cics there is little practical advantage in employing
more refined methods than Algorithm \ref{alg:OptimizingCutoff} for finding an
optimal cutoff $c$ like the hill-climbing approach described in the subsection
"Refined Method For Cutoff Selection" of \cite{Olschewski:2021}.
\par
If the optimal cutoff can not be computed for the whole set of objects because
only some subset of all objects is known to the user then the cutoff
optimization has to be performed for the known set of objects first and used
as an estimator for the whole lot of data later.
\subsubsection{Classification Algorithm}
\label{sssec:ClassificationAlgorithm}
By putting it all together, we arrive at the following algorithm for object
classification.
\\~\\
\begin{algorithm}[H]
  \label{alg:Classification}
  \KwInput{$X\in\MbQ^{m\times n};b^{+},b^{-}\in\MbQ;
      \NumBins\in\MbN_{\geqq3};t^{+},t^{-}\in\MbN_{\geqq1}$}
  \KwOutput{$\Ms{F}(C_{opt})\in\MbB^{|I_{\neg T}|}$}
  \caption{Classification algorithm}
  $\Mr{Scale(X)}$  by
    Algorithm \ref{alg:Scale}\\
    select random subsets $T^{+}\subseteq I^{+}$ of size $t^{+}$,
    $T^{-}\subseteq I^{-}$ of size $t^{-}$\\
  $\Ms{C}:=\Mr{FindCics}(\Mr{Scale}(X),T^{+},T^{-},b^{+},b^{-},\NumBins)$ by
      Algorithm \ref{alg:FindCics} \\
  compute $\Ms{S}_{\Ms{C}}(I_{\neg T})$ by
      Algorithm \ref{alg:ComputingIndicatorValues}\\
  $\Ms{S}_{\Ms{C}}^{\min}:=\min\Ms{S}_{\Ms{C}}(I_{\neg T})$ \\
  $\Ms{S}_{\Ms{C}}^{\max}:=\max\Ms{S}_{\Ms{C}}(I_{\neg T})$ \\
  compute $C_{opt}$ and $\Ms{F}(C_{opt})$ by Algorithm
    \ref{alg:OptimizingCutoff}
\end{algorithm}
~\\
\subsection{Free Parameters:  $b^{+}$, $b^{-}$ And $\NumBins$}
\label{ssec:FreeParameters}
$b^{+}$ and $b^{-}$ specify which features are to be considered as Cics in the
definition of Cic on page \pageref{def:cic}.  They are used in Algorithm
\ref{alg:FindCics}.  In practice, the value of $b^{+}$ seems to be far more
important than $b^{-}$ as most small positive values
$b^{-}\underset{>}{\approx} 0$ served equally well in our applications.
\par
In section \ref{ssec:AutomatizingTheFindingofCics}, we will present Algorithm
\ref{alg:AutoCics} as an alternative for Algorithm \ref{alg:FindCics} in order
to not control the selection of Cics by $b^{+}$ and $b^{-}$ but by specifying
how many of the top ranks according to their $n_{\Mr{diff}}$ value are to be
used as indicator columns.  This replaces $b^{+},b^{-}$---which are obsolete
then---by a new parameter $t$ which in turn may be obsoleted by defining a
convenient default value depending on the feature count of the input data
matrix $X$.
\par
$\NumBins$ specifies how many intervals are to be filled by the histogram
function $H_{a,\NumBins}:\MbQ^{m}\longrightarrow\MbQ^{\NumBins}$ defined
on page \pageref{def:H}.
\par
Note that the output of $\Mr{Scale}(X)$---which serves as input to
$H_{a,\NumBins}$--- may be quantized already, firstly because every continuous
feature of the input data matrix $X\in\MbQ^{m,n}$---measurements of an analog
quantity, for example---is already quantized, and secondly by the well-known
limitations of floating point processing and the number format of $X$ and
$\Mr{Scale}(X)$.  As is typical of algorithms which are processing real-world
data using floating point, quantizations may be obvious (number format of
numerical data in files, for example) or silent (pre-quantized measurement
data, limitations of floating point processing).  In our chip classifications,
we limited the entries of $\Mr{Scale}(X)$ to 3 significant digits with
exponent notation if necessary and set $\NumBins=1000$ or $2000$.
\subsection{Considerations on $\mu_{j}$ And $\sigma_{j}$}
\label{ssec:ConsiderationsOnMuAndSigma}
In practical applications, the means $\mu_{j}$ and the standard deviations
$\sigma_{j}$ ($j=1,\ldots,n$) of the total population as they are needed in
the scaling Algorithm \ref{alg:Scale} may not be known in advance. We can
obtain canonical estimations $\bar{\mu}_{j}$ and $\bar{\sigma}_{j}$ by
computing the sample mean and sample variance of the training objects, i.e.
\begin{eqnarray*}
  \bar\mu_{j} &=& \frac{1}{n}\cdot\sum_{i\in I_{T}}x_{i,j} \\
  s^{2}_{j} &=& \frac{1}{n-1}\cdot\sum_{i\in I_{T}}(x_{i,j}-\mu_{j})^{2}
\end{eqnarray*}
where $I_{T}=T^{+}\cup T^{-}$.  It is well-known that $\bar\mu_{j}$ is an
unbiased estimator for $\mu_{j}$ and $s^{2}_{j}$ is an unbiased estimator for
the variance $\sigma_{j}^{2}$.  For concave functions $f$ of some random
variable $X$ it holds $E[f(X)]\leqq f(E[X])$ but not necessarily equality.
Since the square root is a concave function, $s_{j}:=\sqrt{s^{2}_{j}}$ may not
be an unbiased estimator for $\sigma_{j}$ and generally, there is no formula
for unbiased $\sigma_{j}$ estimation being true for all distributions.  In
case the bias of the estimation of $\sigma_{j}$ being especially low is of
elevated importance there are improved estimators.  See, for example,
\cite{Brugger:1969} and the references therein on using the estimator
$\sqrt{\frac{1}{n-1.5}\cdot\sum\limits_{i\in I_{T}}(x_{i,j}-\mu_{j})^{2}}$ in
case of normal distribution.
\par
Of course, there may be good reasons for decoupling the sample set for
estimating $\mu_{j}$ and $\sigma_{j}$ from $T^{+}$ and $T^{-}$ by choosing a
different set than $I_{T}$ for the former because the size of $I_{T}$ may be
influenced by considerations of avoiding over- and undersampling in the
classification.
\par
\Tb{Note.} It proved useful in our applications to have the possibility to
optionally bypass the $\Mr{FindCics}$ stage (Algorithm \ref{alg:FindCics}) by
implementing a means to specifiy the Cics manually, especially in case of
relevant Cics being known in advance from former runs with similar data.  Then
the $\Ms{C}$ in $\Ms{S}_{\Ms{C}}$ is not the output of $\Mr{FindCics}(\ldots)$
but a list of triples $(j_{s},a^{\ast}_{k_{s}},a^{\ast}_{k_{s}+1})$ created
for those columns $j_{s}$ specified by the user.  This has been applied
successfully in cases where several lots of measurement data are to be
classified whereby the chips are of the same product type.
\subsection{Limits}
\label{ssec:Limits}
Clearly, the classification Algorithm \ref{alg:Classification} requires the
existence of some columns of $X$ where the conditional distribution on the
positive samples differs from the conditional distribution on the negative
samples.  In a stochastic sense, this presupposes the existence of some
features the true distributions of which differ when restricted to the
positive and to the negative objects separately.

\section{Experimental Setup}
\label{sec:ExperimentalSetup}
\subsection{Task And Data}
\label{ssec:TaskAndData}
The tasks to be solved are as described in \cite{Olschewski:2021}:
\Bc
\parbox[t]{0.9\textwidth} {\Em{Given measurements
    and the defect states for chips of a small training set, predict the
    defect state of the remaining chips of the lot based on the measurements
    only.}  }
\Ec
The data we are given consist of measurement data from chip fabrication.  The
data is organized in lots where one lot is a set of wafers from
production. Each wafer carries a fixed number of chips of the same product
type.
\par
In what follows, we ignore the partitioning of a lot into wafers and consider
a lot as a series of $m$ chips of the same product type where each chip is
represented by $n$ measurements. Using the notation of subsection
\ref{ssec:Definitions}, the input to our algorithm consist of:
\begin{itemize}
\item[$-$] $X\in\MbQ^{m\times n}$ where the $i$-th line represents the $i$-th
  chip and the $j$-th column contains the $j$-th measurement
\item[$-$] $v\in\MbB^{n}$ where $v_{i}$ is the true defect state of the $i$-th
  chip.
\end{itemize}
Every column represents a feature because measurements in the same column are
presupposed to be of the same type.
\par
The general hypothesis the classification algorithm presented in
\cite{Olschewski:2021} is based on is that defective devices may possess
abnormalities in patterns derived from deviations by thresholding from the
component-wise means of measurement data.  The approach presented in this
paper in turn resides on the hypothesis that there may exist
features---columns of $X$ in the above algorithms---which we call Cics where
defective devices differ from normal devices in their distributional
properties.
\par
The columns of $X$ are grouped into so-called \Tb{MeasSteps} (measurement
steps) named S1, S2, ... for front-end MeasSteps. Every column of $X$ belongs
to one unique MeasStep.  To every chip and every MeasStep there is assigned
either an error code specifying the type of defect which had occurred in this
step, or \Ts{"0"} meaning ``passed''.  In addition to this, if the tasks
consists in predicting backend defects then to every chip there is assigned a
backend error code or \Ts{"BEpass"}.
\par
In the tasks we are describing here, the true defect state $v_{i}\in\MbB^{m}$
of the $i$-th chip may be
\begin{itemize}
\item[$-$]
the frontend defect state ``anything but \Ts{"0"}'' or
\item[$-$]
the backend defect state ``anything but \Ts{"BEpass"}'' in our tasks.  
\end{itemize}
\Tb{Notation.} We will occasionally abbreviate ``frontend'' as \Tb{FE} and
``backend'' as \Tb{BE}.
\par
The research problem to be solved is: Find algorithms that are able to learn
the function $f:\MbQ^{n}\longrightarrow\MbB$ which maps the vector of
measurements of the $i$-th chip to the defect state of this chip:
\[ (x_{i,1},\ldots,x_{i,n})\mapsto v_{i}
\]
So, given $(x_{i,1},\ldots,x_{i,n})$ for a small set of training chips with
indices $i\in T^{+}\cup T^{-}$ we want an algorithm that predicts
$v_{\bar{i}}$ for chips of the same product type with defect state yet
unknown, knowing only their measurement vectors
$(x_{\bar{i},1},\ldots,x_{\bar{i},n})$.
\par
We are not only interested in predicting defect states of chips when a full
set of measurements is available but also in predictions when only part of
measurement data is available.  This bears the possibilites of saving costs
by reducing the amount of measurements and predicting defects earlier in the
production process.
\par
\subsection{The Data Basis}
\label{ssec:TheDataBasis}
In our experiments we considered chips of 4 different product types A, B, C
and D with the following properties:
\Bc
\begin{tabular}{|l||l|l|l|}
  \hline
  \text{Product} & \#Chips & \#Measurements per chip & continuous only \\
  \hline
  A &  8280 & 385/1084 & yes/no \\
  B &  11952 & 332 & yes \\
  C &  11328 & 915 & no \\
  D &  34550 & 150 & no \\
  \hline
\end{tabular}
\Ec
The data of Product B contains only continuous features (currents,
voltages, ...)  whereas products C and D also include discrete features
(flag words, counts, ...).
\par
An important feature of our method for object classification is that we do not
use any meta-knowledge about the data.  For chip measurement data, this means
that we do not know the types or units nor the meanings of the measurements.
All we know is their numerical values.
\subsection{The Implementation}
\label{ssec:TheImplementation}
A program for the application described in subsection \ref{ssec:TaskAndData}
including several extra features (graphical output etc.) has been realized in
~3400 lines of Python 3 using the \Ts{NumPy}\cite{HarrisEtAl:2020} and
\Ts{Matplotlib}\cite{Hunter:2007} packages.  The algorithms presented above
contain several loops which could be parallelized by using the
\Ts{multiprocessing} package: the first $j$ loop and the $i$ loop in Algorithm
\ref{alg:Scale}, the $j$ loop in Algorithm \ref{alg:FindCics}, the $i$ loop in
Algorithm \ref{alg:ComputingIndicatorValues} and the $c$ loop in
\ref{alg:OptimizingCutoff}.  We keep this optimization for future versions of
this implementation.

\section{Results}
\label{sec:Results}
\subsection{Properties}
\label{ssec:Properties}
The main Algorithm \ref{alg:Classification} meets all of the five goals listed
in section \ref{sec:Introduction} which seems to make it well-suited for
detecting or predicting the overall frontend defect states of chips using
measurement data.
\par
When using the classification algorithm presented here together with the
dimensional reduction method
$\text{dim-reduce}\big(\Mr{Scale}(X),\Mr{sharpness}\big)$ described in
subsection ``Dimensional Reduction'' of \cite{Olschewski:2021} as a
preprocessing step, the algorithm presented here computed better results than
the algorithm of \cite{Olschewski:2021} with the dimensional reduction in a
series of classification runs on the same input.  In what follows, we refer to
this dimensional reduction by \Tb{dim-reduce}.  See \cite{Olschewski:2021} for
more details.  See subsection
\ref{sssec:PredictingFrontendDefectStatesKnowingOnlyPart} for numbers.
\par
The hardware demands of the algorithms presented in this paper are low.  All
classifications the results of which are listed in this paper together took 22
minutes on a Pentium i5-750 using a Python implementation of Algorithm
\ref{alg:Classification} without parallelization on task- or thread-level.
\subsection{Notational Remark}
\label{ssec:NotationalRemark}
We will abbreviate ``true positive'' as \Tb{TP}, ``false positive'' as
\Tb{FP}, ``true negative'' as \Tb{TN} and ``false negative'' as \Tb{FN}
henceforth.
\subsection{Predicting or Detecting Frontend Defect Status of Chips}
\label{ssec:PredictingorDetectingFrontendDefectStatusofChips}
\subsubsection{Predicting Frontend Defect Status S2 of Chips}
\label{sssec:PredictingFrontendDefectStatusS2ofChips}
The goal of this application is to predict the frontend overall defect state
\Ts{"0"} of MeasStep S2 if the S1 and S2 measurements are known to the
classification algorithm.  In the following example, we are going to classify
11328 chips of Product C using 20\%/5\% of all positive/negative chips for
training, setting $b^{+}=0.3$.  The prediction quality is summarized by the 4
fields and their various quotients in the following table:
\begin{center}
\begin{tabular}{|l||r|}
  \hline
  TP &   477   \\
  \hline
  FP &     2   \\
  \hline
  TN & 10175   \\
  \hline
  FN &    15   \\
  \hline
  \hline
  TP/(TP+FN)\% &  97.0 \\
  \hline
  TN/(TN+FP)\% & 100.0 \\
  \hline
  FP/(TP+FN)\% &   0.4 \\
  \hline
  FN/(TN+FP)\% &   0.1 \\
  \hline
  \hline
  TP/FP        & 238.5 \\
  \hline
  TN/FN        & 678.3 \\
  \hline
  \hline
  Accuracy\%   &  99.8 \\
  \hline
  Kappa        & 0.982 \\
  \hline
\end{tabular}
\end{center}
\par
\begin{figure}[ht]
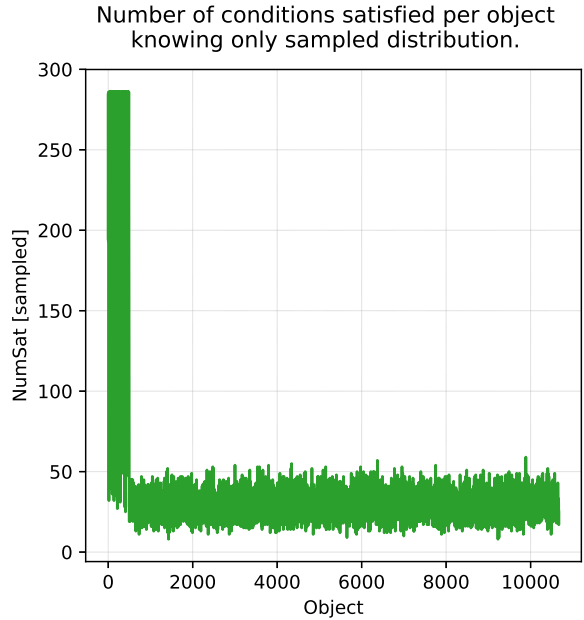

  \caption{$\Ms{S}_{\Ms{C}}(i)$ in the S2 classification with
    product C.}
  \label{fig:1-to-11328-PC-q0q-bp-0-3-bn-0-01-tp-20-tn-5-b-2000-s-1-Sats}
  \centering
  \usegraphics{270}{0}{0.7\textwidth}{0.7\textwidth}{1-to-11328-PC-q0q-bp-0-3-bn-0-01-tp-20-tn-5-b-2000-s-1-Sats}
\end{figure}

The next two figures are created by the Algorithms
\ref{alg:ComputingIndicatorValues} and \ref{alg:OptimizingCutoff} while
solving the aforementioned task.  Figure
\ref{fig:1-to-11328-PC-q0q-bp-0-3-bn-0-01-tp-20-tn-5-b-2000-s-1-Sats} shows
the value $\Ms{S}_{\Ms{C}}(i)$ for each of the 10669 objects $i$ to be
classified.
\par
\Tb{Note.} For the beholder's convience, all positive objects $i$ are
rearranged to the left in this diagram whereas all negative objects $i$ are
moved to the right.  So the differences in $\Ms{S}_{\Ms{C}}(i)$ of positive
objects and negative objects are clearly visible.  This will be done in all
diagrams of this type from now on.
Figure
\ref{fig:1-to-11328-PC-q0q-bp-0-3-bn-0-01-tp-20-tn-5-b-2000-s-1-Cutoffs} shows
the plots of cutoff-vs-accuracy and cutoff-vs-kappa which are created by
Algorithm \ref{alg:OptimizingCutoff} while finding some cutoff $C_{opt}$ for
optimizing accuracy or kappa.
\begin{figure}[ht]
  \caption{Cutoff optimization of the S2 classification with product C.}
  \label{fig:1-to-11328-PC-q0q-bp-0-3-bn-0-01-tp-20-tn-5-b-2000-s-1-Cutoffs}
  \centering
  \usegraphics{270}{0}{0.95\textwidth}{0.5\textwidth}{1-to-11328-PC-q0q-bp-0-3-bn-0-01-tp-20-tn-5-b-2000-s-1-Cutoffs}
\end{figure}
The kappa value of this result is slightly better than the value (0.979)
computed by the algorithm of \cite{Olschewski:2021}.  All in all, this task is
easy to both algorithms even though they differ vastly in the function
principles they are based on.
\par
As another illustration of the performance of Algorithm
\ref{alg:Classification} at predicting frontend defects, we will show next the
results of classifying 34550 chips of product D predicting the overall defect
state \Ts{"0"} of the second MeasStep S2.  We used 50\%/10\% of all
positive/negative objects as samples for training.
\begin{center}
\begin{tabular}{|l||r|}
  \hline
  TP &   740   \\
  \hline
  FP &     0   \\
  \hline
  TN & 31347   \\
  \hline
  FN &    36   \\
  \hline
  \hline
  TP/(TP+FN)\% &  95.4 \\
  \hline
  TN/(TN+FP)\% & 100.0 \\
  \hline
  FP/(TP+FN)\% &   0.0 \\
  \hline
  FN/(TN+FP)\% &   0.1 \\
  \hline
  \hline
  TP/FP        & $\infty$ \\
  \hline
  TN/FN        & 870.8 \\
  \hline
  \hline
  Accuracy\%   &  99.9 \\
  \hline
  Kappa        & 0.976 \\
  \hline
\end{tabular}
\end{center}
\par
The latter result has been achieved by Algorithm \ref{alg:Classification}
using only 3 columns as Cics.  We used the same 3 columns here as we did in
our experiment for minimizing the number of input features for BE defect
prediction, set out in subsection
\ref{sssec:PredictingBackendDefectStatesKnowingOnlyPart}.  When using only 2
columns for Cics or specifying the Cics by $b^{+}$, the resulting quality of
predictions was only marginally different.
\par
The limits mentioned in subsection \ref{ssec:Limits} of the distributional
analysis approach of Algorithm \ref{alg:Classification} showed up when
classifiying Product A.  Unlike with the algorithm of \cite{Olschewski:2021}
where 5000 chips could be classified with a kappa value of 0.707 and accuracy
98.3\%, using Algorithm \ref{alg:Classification} the best result with the same
positive/negative training set sizes 70\%/10\% was kappa 0.469 and accuracy
96.3\%.  In accordance with this, looking at the histograms of Product A, one
sees that when leaving out all discrete features, the positive and negative
histograms of many features actually do not seem to differ much in a way that
histogram analysis could detect with some certainty.  Classifying Product A
was also somewhat harder than classifying the other products mentioned there
to the classification algorithm of \cite{Olschewski:2021}.
\subsubsection{Detecting Frontend Defect Status S1 of Chips}
\label{sssec:DetectingFrontendDefectStatusS1ofChips}
In this application, the task was detecting the frontend defect status S1 of
10000 chips of product B while knowing measurements of S1 and S2.  In order to
know how the detection quality depends on the training set sizes, we tested a
wide range of sizes.  In the following table we list a small representative
selection of results. We set $\NumBins$ to 2000 and $b^{+}$ to 0.98 throughout
this series.
\begin{center}
\begin{tabular}{|r|r||r|r||r|}
  \hline
  \multicolumn{1}{|c|}{Train} & \multicolumn{1}{|c|}{Train}
    & \multicolumn{1}{||c|}{FP} & \multicolumn{1}{|c|}{FN}
    & \multicolumn{1}{||c|}{Kappa} \\
  \multicolumn{1}{|c|}{Pos\%} & \multicolumn{1}{|c|}{Neg\%}
    & \multicolumn{1}{||c|}{} & \multicolumn{1}{|c|}{}
    & \multicolumn{1}{||c|}{} \\
  \hline
       90 & 10 &  0 & 3 & 0.979 \\
       50 & 10 &  1 & 10 & 0.983 \\
       20 & 10 &  1 & 14 & 0.985 \\
        1 & 10 & 16 & 21 & 0.970 \\
  \hline
    0.125 & 10     &  16 & 21 & 0.971 \\
    0.125 &  1     &   5 & 22 & 0.979 \\
    0.125 &  0.125 &   0 & 23 & 0.982 \\
    0.125 &  0.01  &  10 & 12 & 0.983 \\
    0.125 &  0.005 &   1 & 23 & 0.981 \\
  \hline
    0.1   & 10     & 278 & 91 & 0.741 \\
  \hline
\end{tabular}
\end{center}
This table shows that the detection quality is hardly sensitive to the
selection of the training set sizes in this series.  Remarkably, when setting
TrainPos=0.125\% and TrainNeg=0.005\%, the training sets consist of only 3
chips: 2 positive samples and 1 negative sample.
\par
Figure
\ref{fig:1-to-10k-PB-q0q-bp-0-98-bn-0-01-tp-0-125-tn-0-005-b-2000-s-1-Cutoffs}
shows the plots of cutoff-vs-accuracy and cutoff-vs-kappa which are created by
Algorithm \ref{alg:OptimizingCutoff} while classifying 10000 chips of product
B using 2 positive sample chips and 1 negative sample chip for training.
\begin{figure}[ht]
  \caption{Cutoff optimization of the S1 classification with product B.}
  \label{fig:1-to-10k-PB-q0q-bp-0-98-bn-0-01-tp-0-125-tn-0-005-b-2000-s-1-Cutoffs}
  \centering
  \usegraphics{270}{0}{0.95\textwidth}{0.5\textwidth}{1-to-10k-PB-q0q-bp-0-98-bn-0-01-tp-0-125-tn-0-005-b-2000-s-1-Cutoffs}
\end{figure}
\par
\begin{figure}[ht]
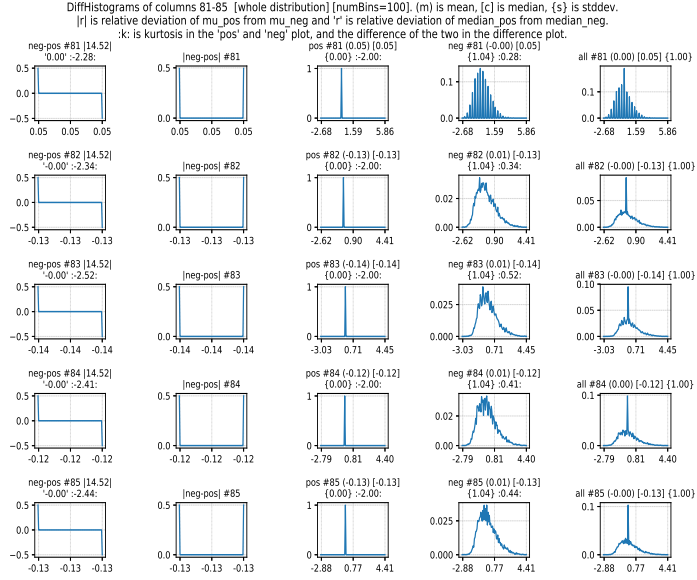

  \caption{Histogram collection of 5 features, knowing all objects.}
  \label{fig:Prod-B-Histo-2-2-Train-Full}
  \centering
  \usegraphics{270}{0}{0.9\textwidth}{0.7\textwidth}{Prod-B-Histo-2-2-Train-Full}
\end{figure}

The next 3 figures show some histograms of the above series, created by
Algorithm \ref{alg:FindCics}.  In each of the 3 figures, there are 25 diagrams
in a $5\times5$ grid.  The five rows correspond to columns \#81 to \#85 of the
matrix $\Mr{Scale}(X)$.  There are 5 histograms in each row, each rendered by
using 100 bins:
\begin{enumerate}
  \item the (signed) difference of the pos and neg histograms
  \item their absolute difference
  \item the histogram of positive objects (``pos'')
  \item the histogram of negative objects (``neg'')
  \item the histogram of all objects (``all'')
\end{enumerate}
Figure \ref{fig:Prod-B-Histo-2-2-Train-Full} shows how the histograms would be
looking like if the classification algorithm would get to see all objects as
input data.  In real applications the algorithms gets to see only the training
sets.  So in the next Figure \ref{fig:Prod-B-Histo-NormalTrain-Samp} the
histograms are displayed when using 20\%/10\% of all objects as samples for
the positive/negative training sets.  The topmost row displays column \#81
which is not used as a Cic (see page \pageref{def:cic}) in classification,
whereas the next 4 rows (displaying columns \#82 to \#85) are used as Cics.
In accordance with this, if we compare the 3rd column of Figure
\ref{fig:Prod-B-Histo-NormalTrain-Samp} with the 4th column, we see that in
the 4th column, the peak of the topmost diagram is much higher than in the
remaining 4 diagrams of this column.  This means, the difference in relative
frequencies of the most frequent bin of the ``pos'' histogram (column 3) and
the ``neg'' histogram (column 4) is much lower in column \#81 than in the
columns \#82 to \#85.  Thus Algorithm \ref{alg:FindCics} will use columns \#82
to \#85 as Cics, but not \#81.
\par
\begin{figure}[ht]
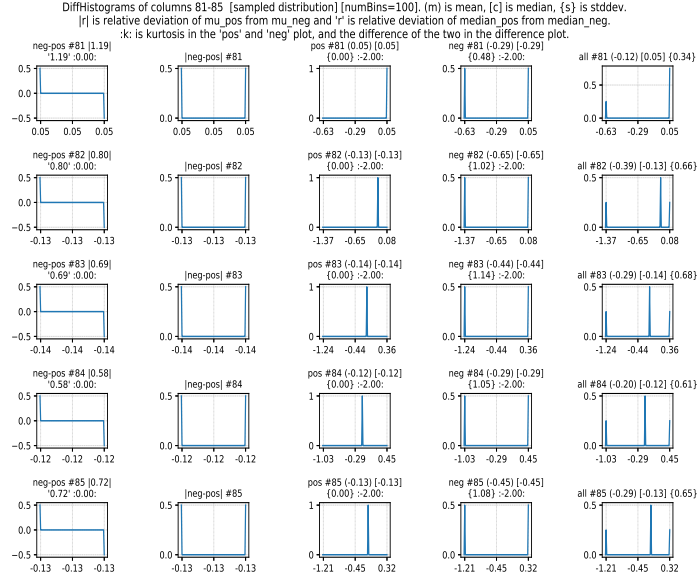

  \caption{Histogram collection of 5 features, knowing 2+2 sample objects only.}
  \label{fig:Prod-B-Histo-2-2-Train-Samp}
  \centering
  \usegraphics{270}{0}{0.9\textwidth}{0.7\textwidth}{Prod-B-Histo-2-2-Train-Samp}
\end{figure}

Finally, in Figure \ref{fig:Prod-B-Histo-2-2-Train-Samp} we see the histograms
created just like the former, but this time using only 4 chips for training: 2
positive objects and 2 negative objects.  As one can see, the only relative
frequencies occurring are $0,0.25,0.5,0.75$ and $1$. Again, by comparing the
diagrams of the 3rd column (``pos'') with the diagrams of the 4th column
(``neg''), we see that the peaks in the 4 bottom ``pos'' histograms are
located differently than in the belonging 4 bottom ``neg'' histograms of the
same row.  But this is not the case when comparing the topmost ``pos'' with
the topmost ``neg'' histogram, which again gives a hint why the FindCics
Algorithm \ref{alg:FindCics} takes columns \#82 to \#85 as Cics, but not \#81,
even when knowing only 2+2 samples for training.
\begin{figure}[ht]
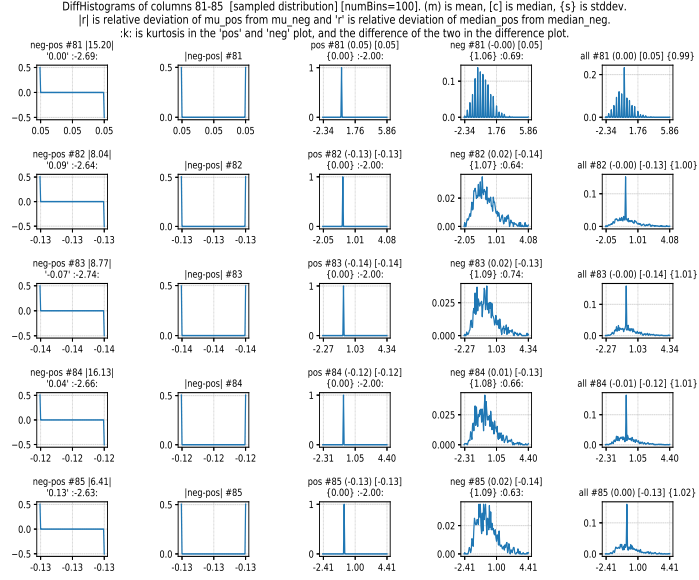

  \caption{Histogram collection of 5 features, knowing 20\%/10\% sample sets only.}
  \label{fig:Prod-B-Histo-NormalTrain-Samp}
  \centering
  \usegraphics{270}{0}{0.9\textwidth}{0.7\textwidth}{Prod-B-Histo-NormalTrain-Samp}
\end{figure}
\subsubsection{Predicting Frontend Defect States Knowing Only Part of Measurement Data}
\label{sssec:PredictingFrontendDefectStatesKnowingOnlyPart}
In order to evaluate the suitability of the algorithm described in this paper
to the prediction of the frontend overall defect state \Ts{"0"} with a
successively reduced set of features as input, we repeated the application
described in section ``Dimensional Reduction'' of \cite{Olschewski:2021} with
the same input data but replaced the algorithm of loc. cit. by the
classification algorithm described in this paper.  
\par
After scaling the input matrix $X$ by $\Mr{Scale}(X)$ in Algorithm
\ref{alg:Classification}, we proceeded by processing the scaled matrix by
running $\text{dim-reduce}\big(\Mr{Scale}(X),\Mr{sharpness}\big)$ with
increasing sharpness settings 0,1,2,3,4. Both algorithms have been extended by
this dim-reduce preprocessing.
\par
The next two tables show the results of classifying 11328 chips of
product C (429 S1-measurements per chip). The algorithms are given
measurements of MeasStep S1 only, and the task is predicting the overall defect
state \Ts{"0"} of MeasStep S2 without knowing any S2 measurements.
\par
The following table shows the quality of prediction by the
classification algorithm of \cite{Olschewski:2021} using abs-t-excess
thresholding.
\Bc
\begin{tabular}{|r|r|r|r|r|r|}
  \hline
  \multicolumn{1}{|c|}{Sharpness} & \multicolumn{1}{|c|}{\#Features}
    & \multicolumn{1}{|c|}{\%Features}
    & \multicolumn{1}{|c|}{Accuracy} & \multicolumn{1}{|c|}{Kappa}
    & \multicolumn{1}{|c|}{\underline{TP}} \\
  \multicolumn{1}{|c|}{of reduction} & \multicolumn{1}{|c|}{omitted}
    & \multicolumn{1}{|c|}{omitted}
    & \multicolumn{1}{|c|}{[$z_{min}$]} & \multicolumn{1}{|c|}{[$z_{min}$]}
    & \multicolumn{1}{|c|}{FP} \\
  \hline
       0 &   0 &    0\% & 0.989 & 0.885 & 33.3 \\
       1 & 283 & 66.0\% & 0.974 & 0.664 & $+\infty$ \\
       2 & 346 & 80.7\% & 0.973 & 0.656 & 34.4 \\
       3 & 379 & 88.3\% & 0.946 & 0.194 & 0.234 \\
       4 & 387 & 90.2\% & 0.946 & 0.188 & 0.201 \\
  \hline
\end{tabular}
\Ec
In this second table we list the corresponding results by the
algorithm described in this paper, using 50\%/5\% of all positive/negative
chips as samples:
\Bc
\begin{tabular}{|r|r|r|r|r|r|}
  \hline
  \multicolumn{1}{|c|}{Sharpness} & \multicolumn{1}{|c|}{\#Features}
    & \multicolumn{1}{|c|}{\%Features}
    & \multicolumn{1}{|c|}{Accuracy} & \multicolumn{1}{|c|}{Kappa}
    & \multicolumn{1}{|c|}{\underline{TP}} \\
  \multicolumn{1}{|c|}{of reduction} & \multicolumn{1}{|c|}{omitted}
    & \multicolumn{1}{|c|}{omitted}
    & \multicolumn{1}{|c|}{} & \multicolumn{1}{|c|}{}
    & \multicolumn{1}{|c|}{FP} \\
  \hline
       0 &   0 &    0\% & 0.997  & 0.936 & 272 \\
       1 & 283 & 66.0\% & 0.996  & 0.931 & 90.3 \\
       2 & 346 & 80.7\% & 0.988  & 0.769 & 9.5 \\
       3 & 379 & 88.3\% & 0.985  & 0.638 & 73.5 \\
       4 & 387 & 90.2\% & 0.856  & 0.135 & 0.1 \\
  \hline
\end{tabular}
\Ec
As for the accuracy and kappa values, the algorithm described in this paper is
superior to the classification algorithm of \cite{Olschewski:2021} in tackling
this task with sharpness 0 to 3.  For example, with sharpness 1---leaving out
66\% of the features---the algorithm of this paper reaches a kappa value of
0.931 which is excellent in comparison to 0.664 by the algorithm of
\cite{Olschewski:2021}.
\par
With sharpness 3---leaving out 88.3\% of the features---the cited algorithm
reaches a rather low kappa value of 0.194 and
$\frac{\Mr{TP}}{\Mr{FP}}<1$ whereas the algorithm of this paper still reaches
0.638 and $\frac{\Mr{TP}}{\Mr{FP}}>73$.
\subsection{Predicting Backend Defect Status of Chips}
\label{ssec:PredictingBackendDefectStatusofChips}
All results described up to now were obtained by trying to detect or to
predict the frontend (FE) overall defect state \Ts{"0"} of chips from a series
of wafers. A related problem is trying to predict the backend (BE) overall
defect state \Ts{"BEpass"} knowing only frontend measurement data.  The latter
problem of BE defect prediction has turned out to be harder than FE defect
prediction or detection to all algorithms we have tried.
\par
\Tb{Note.}  Whenever we write ``predicting the BE overall defect state
\Ts{"BEpass"}'', what we actually are doing is predicting the negation of
\Ts{"BEpass"}.  This is completely equivalent because the predictions are
binary so negating the predicted bit converts the former into the latter
and vice versa.
\subsubsection{Predicting Backend Defect Status of Chips Knowing Frontend Data}
\label{sssec:PredictingBackendDefectStatusofChipsKnowingFrontendData}
In this application, input data consists of 150 measurements per chip taken at
different frontend fabrication steps, and, as in the FE application, all chips
are of the same product type.  The task consists in predicting the BE overall
defect state \Ts{"BEpass"} for a large sets of chips after training with two
smaller sets of positive and negative samples, respectively.  In addition, we
wanted to know how many and which measurements could be omitted without
reducing the quality of prediction unduly.
\par
\begin{figure}[ht]
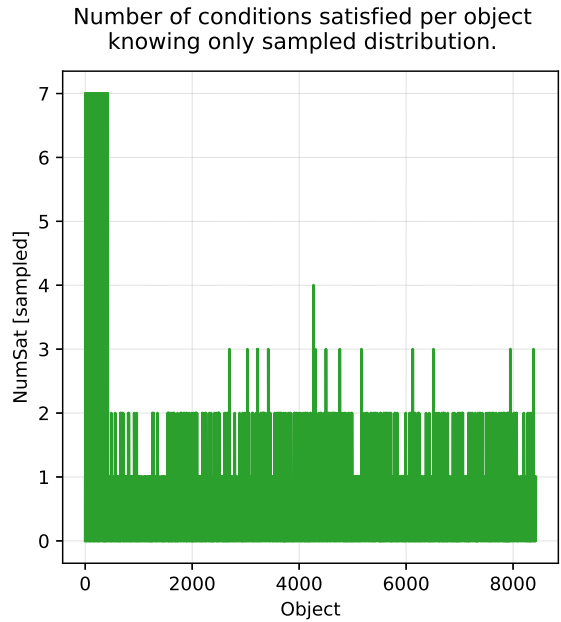

  \caption{$\Ms{S}_{\Ms{C}}(i)$ in BE classification with product D, using
    all features.}
  \label{fig:1-to-10k-PD-qBEpassq-bp-0-41-bn-0-01-tp-60-tn-10-b-2000-s-1-Sats}
  \centering
  \usegraphics{270}{0}{0.7\textwidth}{0.7\textwidth}{1-to-10k-PD-qBEpassq-bp-0-41-bn-0-01-tp-60-tn-10-b-2000-s-1-Sats}
\end{figure}
\begin{figure}[ht]
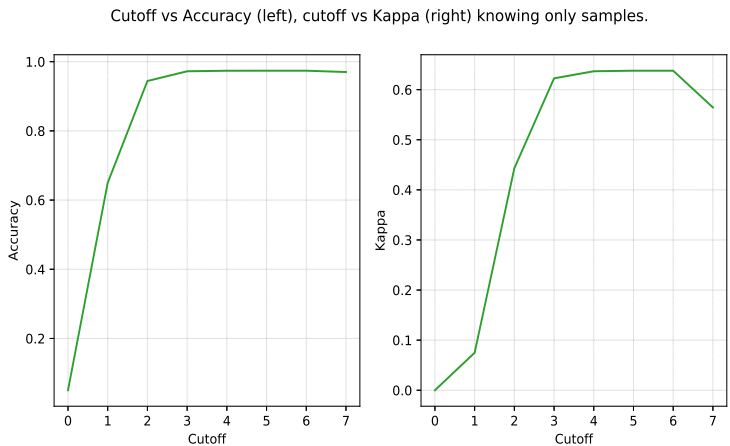

  \caption{Cutoff optimization of BE classification with product D, using
    all features.}
  \label{fig:1-to-10k-PD-qBEpassq-bp-0-41-bn-0-01-tp-60-tn-10-b-2000-s-1-Cutoffs}
  \centering
  \usegraphics{270}{0}{0.95\textwidth}{0.5\textwidth}{1-to-10k-PD-qBEpassq-bp-0-41-bn-0-01-tp-60-tn-10-b-2000-s-1-Cutoffs}
\end{figure}

In the following two figures
\ref{fig:1-to-10k-PD-qBEpassq-bp-0-41-bn-0-01-tp-60-tn-10-b-2000-s-1-Sats}
and
\ref{fig:1-to-10k-PD-qBEpassq-bp-0-41-bn-0-01-tp-60-tn-10-b-2000-s-1-Cutoffs},
we see plots created by Algorithm \ref{alg:Classification} while classifying
10000 chips of product D for BE overall defect status. See the explanations
for Figure
\ref{fig:1-to-11328-PC-q0q-bp-0-3-bn-0-01-tp-20-tn-5-b-2000-s-1-Cutoffs} and
\ref{fig:1-to-11328-PC-q0q-bp-0-3-bn-0-01-tp-20-tn-5-b-2000-s-1-Sats} in
subsection \ref{sssec:PredictingFrontendDefectStatusS2ofChips} for details on
the type of results these plots show.  For this BE classification of 10000
chips of product D we set $b^{+}=0.41$ and used 60\% of all positive and 10\%
of all negative chips as samples for training.
\subsubsection{Predicting Backend Defect States Knowing Only Part of Measurement Data}
\label{sssec:PredictingBackendDefectStatesKnowingOnlyPart}
Next, we tried how far we can reduce the number of features without inducing
inacceptable loss in quality of prediction.  In result, we could shrink the
input data to just 2 specific features out of 150.  We used the same data lot
and the same amount of positive (60\%) and negative (10\%) chips for training
as in subsection
\ref{sssec:PredictingBackendDefectStatusofChipsKnowingFrontendData} above.
The plots in Figure
\ref{fig:1-to-10k-PD-qBEpassq-bp-0-2-bn-0-01-tp-60-tn-10-b-2000-s-1-i142i144-Sats}
and
\ref{fig:1-to-10k-PD-qBEpassq-bp-0-2-bn-0-01-tp-60-tn-10-b-2000-s-1-i142i144-Cutoffs}
are created by Algorithm \ref{alg:Classification} using only these 2 features.
\begin{figure}[ht]
  \caption{$\Ms{S}_{\Ms{C}}(i)$ in BE classification with product D,
    using only 2 features.}
  \label{fig:1-to-10k-PD-qBEpassq-bp-0-2-bn-0-01-tp-60-tn-10-b-2000-s-1-i142i144-Sats}
  \centering
  \usegraphics{270}{0}{0.7\textwidth}{0.7\textwidth}{1-to-10k-PD-qBEpassq-bp-0-2-bn-0-01-tp-60-tn-10-b-2000-s-1-i142i144-Sats}
\end{figure}
\begin{figure}[ht]
  \caption{Cutoff optimization of BE classification with product D, using
    only 2 features.}
  \label{fig:1-to-10k-PD-qBEpassq-bp-0-2-bn-0-01-tp-60-tn-10-b-2000-s-1-i142i144-Cutoffs}
  \centering
  \usegraphics{270}{0}{0.95\textwidth}{0.5\textwidth}{1-to-10k-PD-qBEpassq-bp-0-2-bn-0-01-tp-60-tn-10-b-2000-s-1-i142i144-Cutoffs}
\end{figure}
The following table shows example results for the aforementioned BE
classification of 10000 chips of product D, using all 150 features or using
just 2 or 3 specific ones.
\begin{center}
\begin{tabular}{|r|r||r|r||r|r|}
  \hline
  \multicolumn{1}{|c|}{\#Features} & \multicolumn{1}{|c|}{\#Cics}
    & \multicolumn{1}{||c|}{TP/FP} & \multicolumn{1}{|c|}{TN/FN}
    & \multicolumn{1}{||c|}{Accuracy\%}
    & \multicolumn{1}{|c|}{Kappa} \\
  \hline
      150 &  7 &  $\infty$ & 36.4 & 97.4 & 0.638 \\
  \hline
        3 &  3 &  $\infty$ & 36.4 & 97.4 & 0.638 \\
  \hline
        2 &  2 &       204 & 36.4 & 97.4 & 0.637 \\
  \hline
\end{tabular}
\end{center}
As can be seen in this table, if using all 150 columns, Algorithm
\ref{alg:FindCics} finds 7 Cics.  When we replaced this by 3 or even 2
very specific columns, the prediction quality by Algorithm
\ref{alg:Classification} was only marginally worse.
\subsubsection{Addendum: Batch-Wise Cutoff Optimization}
\label{sssec:BatchWiseCutoffOptimization}
If measurement data of several wafers of the same product type is put together
to some data lot, inhomogenities may occur, for example caused by parameter
shifts or by accumulation of defects in certain wafers.  As a consequence,
optimizing the cutoff over a complete data lot containing the measurement data
of---for example---some dozens of wafers, as Algorithm
\ref{alg:OptimizingCutoff} does by default, may be too coarse.  In order to
explore this further, we performed a series of experiments by not optimizing
the cutoff once over the whole series of wafers but by repeating Algorithm
\ref{alg:OptimizingCutoff} on a sequence of batches each containing a fixed
number of chips instead.
\par
In the rest of this subsection, results relate to predicting the BE overall
defect state \Ts{"BEpass"} of product D (34550 chips) using 70\%/5\% of all
positive/negative objects as samples for training, using 3 columns out of 150
as Cics as described in subsection
\ref{sssec:PredictingBackendDefectStatesKnowingOnlyPart}.  The following table
shows the results of some sequences of runs of the classification Algorithm
\ref{alg:Classification} when performing the optimization step by Algorithm
\ref{alg:OptimizingCutoff} on a single batch instead of the complete lot.
There is considerable variation in the prediction quality for separate
batches, as the following table shows.
\begin{center}
\begin{tabular}{|rl||r|r|r|r|r|}
  \hline
      &       & 200   &   400  &  1000  &  1382 &  4000 \\
  \hline
\hline
  Max & Kappa & 1.0   & 0.920  &  0.729 & 0.669 & 0.619 \\
\hline
  Min & Kappa & 0.0   & 0.196  &  0.433 & 0.423 & 0.535 \\
\hline
\end{tabular}
\end{center}
Using a batch size of 1382 means wafer-wise cutoff optimization.  Looking at
how many batches could be classified with given kappa values, we obtained the
following numbers.
\begin{center}
\begin{tabular}{|l||r|r|r|r|r|}
\hline
  & Batchsize 200 & Batchsize 400 \\
  \hline
  \hline
  \#Batches with $\kappa=1.0$           & 13  &  0 \\
  \hline
  \#Batches with $\kappa\in[0.9,1.0)$   &  1  &  2 \\
  \hline
  \#Batches with $\kappa\in[0.8,0.9)$   & 11  &  2 \\
  \hline
  \#Batches with $\kappa\in[0.7,0.8)$   & 29  & 16 \\
  \hline
\end{tabular}
\end{center}
\subsection{Classifying Iris Types}
\label{ssec:ClassifyingIrisTypes}
As a completely different application in multiple respects---object count,
feature count and type of measurements---we tested Algorithm
\ref{alg:Classification} with the classic iris flower data
set\cite{Fisher:1936} with the correction of \cite{Dua:2019}.  It turns out
that Algorithm \ref{alg:Classification} performs well on this data set despite
of its small number of objects and features.  The following results have all
been obtained using $\NumBins=5$ or $6$, 60\%/1\% positive/negative training
set size and specifying columns \#3 and \#4 as Cics.  The selection of Cics
can also be done by prescribing $b^{+}$ (see Algorithm \ref{alg:FindCics}),
with slightly worse quality of prediction.
\begin{center}
  \begin{tabular}{|c||c|c|c|c||c|c|}
  \hline
  Type & TP & FP & TN & FN & Kappa & Accuracy\% \\
    \hline
    \hline
    setosa     & 18 &  0 & 99 &  2 & 0.937 & 98.3 \\
    \hline
    versicolor & 13 &  1 & 98 &  7 & 0.727 & 93.3 \\
    \hline
    virginica  & 13 &  0 & 99 &  7 & 0.756 & 94.1 \\
    \hline
  \end{tabular}
\end{center}
In comparison, the classification algorithm of \cite{Olschewski:2021}
performed well only when classifying two of the three types.
\subsection{Dimensional Reduction}
\label{ssec:DimensionalReduction}
As rendered by examples in subsections
\ref{sssec:PredictingBackendDefectStatesKnowingOnlyPart} and
\ref{sssec:PredictingFrontendDefectStatusS2ofChips}, there are applications
where not all features are necessary for chip classifications.  Some features
or even groups of features in the baselying measurement data may be eminently
important whereas others may be dispensable without much degradation in
prediction quality.  We will be using the notations of section
\ref{sec:TheAlgorithms} here.
\par
Let $H_{a^{\ast},\NumBins}:\MbQ^{m}\to\MbQ^{\NumBins}$ be the histogram
function mapping $m$ numbers to the relative frequencies
$(h_{0},\ldots,h_{\NumBins-1})$ of these $m$ numbers if distributing them to
$\NumBins$ equidistant intervals $I_{0},\ldots,I_{\NumBins-1}$ with bounds
$a^{\ast}$.
\par
\Tb{Definitions.} Let $j$ be some column index in $\{1,\ldots,n\}$.  Consider
$(h_{0},\ldots,h_{\NumBins-1})=H_{a^{\ast},\NumBins}(x_{+,j})$ where
$x_{+,j}=\left(x_{i,j}:i\in T^{+}\right)\in\MbQ^{|T^{+}|}$ is the $j$-th
column of $\Mr{Scale}(X)$, reduced to those rows representing positive
training objects.  Let $k^{\ast}$ be the smallest index in
$\{0,\ldots,\NumBins-1\}$ with
$h_{k^{\ast}}=\max(h_{0},\ldots,h_{\NumBins-1})$.  Then $I_{k^{\ast}}$ is the
(left-most) interval of those occurring with maximum relative frequency.
Define
\begin{eqnarray*}
  n_{\Mr{pos}}(j) &=& |\{ i\in T^{+}:x_{i,j}\in I_{k^{\ast}}\}| \\
  n_{\Mr{neg}}(j) &=& |\{ i\in T^{-}:x_{i,j}\in I_{k^{\ast}}\}| \\
  n_{\Mr{diff}}(j) &=& n_{\Mr{pos}}(j) - n_{\Mr{neg}}(j)\\
\end{eqnarray*}
\par
As follows from the definition of Cic in subsection
\ref{sssec:FindingCandidateIndicatorColumns}, some column $j$ is suited for
being used as one Cic in computing the indicators by Algorithm
\ref{alg:ComputingIndicatorValues} if the property of $x_{i,j}$ lying in the
most-frequent interval of the $j$-th column's histogram can be used as a
criterion separating as many as possible positive objects
$(x_{i,1},\ldots,x_{i,n})$ from negative objects.  Since all the
classification algorithm knows in training is the samples, we must take
histograms of the columns reduced to the positive objects,
$(h_{0},\ldots,h_{\NumBins-1})=H_{a^{\ast},\NumBins}(x_{+,j})$ as defined
above.
\par
So some column $j$ is suited for being used as one Cic in computing the
indicators by Algorithm \ref{alg:ComputingIndicatorValues} if it has the
following two properties.
\begin{itemize}
\item[-] For many positive samples $(x_{i,1},\ldots,x_{i,n})$ ($i\in T^{+}$),
  $x_{i,j}$ lies in the most-frequent interval $I_{k^{\ast}}$ of
  $H_{a^{\ast},\NumBins}(x_{+},j)$.
\item[-] For few negative samples $(x_{\bar{\i},1},\ldots,x_{\bar{\i},n})$
  ($\bar{\i}\in T^{-}$), $x_{\bar{\i},j}$ lies in the aforementioned
  $I_{k^{\ast}}$.
\end{itemize}
Therefore, those columns $j$ that satisfy these two criteria to a high extent
possess high $n_{\Mr{diff}}(j) = n_{\Mr{pos}}(j) - n_{\Mr{neg}}(j)$ values
which suggests using $n_{\Mr{diff}}(j)$ as a relevance indicator of column
$j$.
\par
As an example, here is the top of some table listing column indices $j$
ordered by decreasing $n_{\Mr{diff}}(j)$:
\begin{center}
  \begin{tabular}{|c||c||c|c|c|}
\hline
 Rank &  $j$  & $n_{\Mr{diff}}$ & $n_{\Mr{pos}}$ & $n_{\Mr{neg}}$ \\
\hline
 1  & 144  &     844  &     879  &      35 \\
 2  & 145  &     841  &     899  &      58 \\
 3  & 142  &     836  &     903  &      67 \\
 4  & 141  &     831  &     897  &      66 \\
 5  & 143  &     825  &     901  &      76 \\
\hline   
   $\cdots$ & $\cdots$ & $\cdots$ & $\cdots$ & $\cdots$ \\
\hline   
 146  &  1  &       0  &      24  &      24 \\
 147  &  5  &     -11  &      64  &      75 \\
 148  & 17  &    -112  &     271  &     383 \\
 149  & 16  &    -130  &     346  &     476 \\
 150  &  6  &    -230  &     765  &     995 \\
\hline
\end{tabular}
\end{center}
This table has been extracted while classifying product D for the BE overall
defect state as described in
\ref{sssec:PredictingBackendDefectStatesKnowingOnlyPart}.  As is clearly
visible, the 3 columns $j=142,143,144$ occur among the top five ranks.  These
3 columns were those that made it possible to use only 3 out of 150 columns as
Cics with a negligible degradation of prediction quality in FE
prediction---see subsection
\ref{sssec:PredictingFrontendDefectStatusS2ofChips} ---and in BE
prediction---see subsection
\ref{sssec:PredictingBackendDefectStatesKnowingOnlyPart}.
\subsection{Automatizing The Finding of Cics}
\label{ssec:AutomatizingTheFindingofCics}
The $n_{\Mr{diff}}$ ranking detailed in section
\ref{ssec:DimensionalReduction} can be used in order to try finding relevant
features, i.e., features that should be taken into respect when computing the
indicator values $\Ts{S}_{\Ts{C}}(i)$ by Algorithm
\ref{alg:ComputingIndicatorValues}.  The resulting classification algorithm is
the same as Algorithm \ref{alg:Classification} with the only difference being
that the step
\begin{center}
$\Ms{C}:=\Mr{FindCics}(\Mr{Scale}(X),T^{+},T^{-},b^{+},b^{-},\NumBins)$ by
Algorithm \ref{alg:FindCics}
\end{center}
is replaced by
\begin{center}
$\Ms{C}:=\Mr{AutoCics}(\Mr{Scale}(X),T^{+},T^{-},\NumBins,t)$ by
Algorithm \ref{alg:AutoCics}
\end{center}
which refers to the following algorithm.  Let $t$ be some number in
$\{1,2,\ldots,n\}$ where $n$ is the number of columns of the input data matrix
$X$.
\\~\\
\begin{algorithm}[H]
  \label{alg:AutoCics}
  \KwInput{$\Mr{Scale}(X)\in\MbQ^{m\times n},T^{+}\subseteq I^{+},
    T^{-}\subseteq I^{-},\NumBins\in\MbN_{\geqq3},t\in\{1,\ldots,n\}$}
  \KwOutput{list $L=\left[(j_{s},a^{\ast}_{k_{s}},a^{\ast}_{k_{s}+1})
    \mid s=1,\ldots,t\right]$}
  \caption{AutoCics($X,T^{+},T^{-},\NumBins,t$)}
  \For{$j=1\ldots n$}
  {
    set $x^{+}=(x_{i,j}\mid i\in T^{+})$ \\
    compute $a^{\ast}=a^{\ast}_{\NumBins}(x^{+})$ \\
    compute $n_{\Mr{diff}}(j)$ as in section \ref{ssec:DimensionalReduction} 
  }
  $R=\left[\big(j,n_{\Mr{diff}}(j)\big)\mid j=1,\ldots,n\text{; ordered by
      decreasing $n_{\Mr{diff}}(j)$}\right]$ \\
  set $\left[ \big(j_{1},n_{\Mr{diff}}(j_{1})\big),
     \ldots,\big(j_{n},n_{\Mr{diff}}(j_{n})\big)\right]=R$ \\
  \Return $L=\left[(j_{s},a^{\ast}_{k_{s}},a^{\ast}_{k_{s}+1})
    \mid s=1,\ldots,t\right]$ 
\end{algorithm}
~\\
If $t$ is set to some number in $\{1,\ldots,n\}$ then Algorithm
\ref{alg:AutoCics} returns this many columns with the highest $n_{\Mr{diff}}$
values and thus can be plugged into the classification algorithm
\ref{alg:Classification} as an alternative for computing $\Ms{C}$ by Algorithm
\ref{alg:FindCics}.
\subsubsection{Practical Results by Using AutoCics()}
\label{sssec:PracticalResultsByUsingAutoCics}
In order to get an impression of the practicability of the AutoCics() variant
of Algorithm \ref{alg:Classification}, we tested the degree of stability of
the column list $L$ computed by the AutoCics() Algorithm \ref{alg:AutoCics}
against changing training set selection, training set size and changing data
lots of the same product type. 
\par
In order to compare the AutoCics() variant with results mentioned earlier in
this paper, we repeated the task as described in section
\ref{sssec:DetectingFrontendDefectStatusS1ofChips}: predict the S1 overall
defect state of 10000 chips of Product B using minimal training sets of only 2
positive and 1 negative samples.  But unlike
\ref{sssec:DetectingFrontendDefectStatusS1ofChips}, we do not control the
selection of Cics by $b^{+}$ but by the parameter $t$ of the AutoCics()
algorithm. $b^{+}$ and $b^{-}$ are not used here.
\par
The following table shows results for three different settings of the
parameter $t$---called ``\#Top Cics Used'' in this table---in Algorithm
\ref{alg:AutoCics}.  When setting $t=50$, the difference in detection quality
is small with Kappa changing from 0.983 to 0.980 and FP+FN increasing from
10+12 to 2+23.  Because of these tiny training sets, only $50\cdot 3$ numbers
of the matrix $\Mr{Scale(X)}$ are used for classifying 10000 chips here.
\begin{center}
\begin{tabular}{|c||r|r||r|r|}
  \hline
  {\#Top} & {FP} & {FN} & {Accuracy\%} & {Kappa} \\
  {Cics Used} & {} & {} & {} & {} \\
  \hline
       50 & 2 &  23 & 99.7 & 0.980 \\
       45 & 81 &  23 & 99.0 & 0.921 \\
       40 & 277 &  23 & 97.0 & 0.799 \\
  \hline
\end{tabular}
\end{center}
\subsubsection{Obsoleting Free Parameters by Using AutoCics()}
\label{sssec:ObsoletingFreeParametersbyUsingAutoCics}
Using AutoCics() by Algorithm \ref{alg:AutoCics} offers the possibility to get
rid of the necessity of controlling the selection of Cics by the user.  We can
specify a default value $t^{\ast}$ for $t$, derived from $n$ (= number of
columns of $X$), which is to be used if the user does not specify $t$ himself.
This makes it possible to do without specifying $b^{+},b^{-}$ in Algorithm
\ref{alg:Classification} as well as $t$ in Algorithm \ref{alg:AutoCics}.
\par
We set $t^{\ast}=\lceil 0.1\cdot n\rceil$ so that the top 10\% ranks of the
$n_{\Mr{diff}}$ table described in sections \ref{ssec:DimensionalReduction}
and \ref{ssec:AutomatizingTheFindingofCics} are used as indicator columns.
\par
The following table shows results of the setting described in
\ref{sssec:PracticalResultsByUsingAutoCics}, but this time $t^{\ast}$ is used
for $t$.  As Product B contains 332 measurements per chip, $t^{\ast}$ is 34
here.  We use 5 chips for training, 3 positive samples and 2 negative ones.
\begin{center}
\begin{tabular}{|l||r|}
  \hline
  TP &   657   \\
  FP &    13   \\
  TN &  9301   \\
  FN &    25   \\
  \hline
  TP/(TP+FN)\% &  96.3 \\
  TN/(TN+FP)\% &  99.9 \\
  FP/(TP+FN)\% &   1.9 \\
  FN/(TN+FP)\% &   0.3 \\
  \hline
  TP/FP        &  50.5 \\
  TN/FN        & 372.0 \\
  \hline
  Accuracy\%   &  99.6 \\
  Kappa        & 0.970 \\
  \hline
\end{tabular}
\end{center}

\section{Algorithmic Complexity}
\label{sec:AlgorithmicComplexity}
In what follows, we count operations $+,-,\times,\div$, $\leqq$,
$\lfloor\cdot\rfloor$ and $\sqrt{\cdot}$ at unit cost for all occurring
numbers.  This is a realistic model when using fixed-width rational number
types like 64-bit floating point on a real computer.  By this, we assume a
somewhat idealized computer by ignoring certain limitations of data types
frequently used for unit-cost- (or constant-cost)-arithmetics on real computer
hardware: limited value range, rounding errors in representing rational
numbers whose denominator is not a power of $2$, approximations of
$\sqrt{\cdot}$ etc.  Furthermore, we ignore costs associated with accessing
and modifying simple data structures---mainly, accessing single numbers,
vectors and matrices of numbers in memory.  Counting these may increase costs
by polylogarithmic factors, depending on the chosen data structures, the input
size and the computer hardware the user is going to use.
\subsection{Scaling (Algorithm \ref{alg:Scale})}
For $X\in\MbQ^{m\times n}$ we need to compute $n$ means and $n$ standard
deviations in $\McO(m\cdot n)$, and compute
$\frac{x_{i,j}-\mu_{j}}{\sigma_{j}}$ for all $i,j$. Altogether, this takes
\[  \McO(m\cdot n)
\]
arithmetic operations, including $n$ square roots.
\subsection{Computing one histogram
    $\Mr{H}_{a,\NumBins}(x_{1},\ldots,x_{\bar{m}})$}
Let $a=(-\infty<a_{1}<\cdots<a_{\NumBins-1}<\infty)$.
\par
\Tb{Case 1:} The inner boundaries $a_{1},\ldots,a_{\NumBins-1}$ are chosen by
the user: For every $i$ we find the unique bin $[a_{k},a_{k+1})$ which
contains $x_{i}$ by binary search on the boundaries vector $a$, which takes
$\log(\NumBins)$ comparisons plus incrementing one of $\NumBins$ counters.  So
the overall complexity is in
\[  \McO\big(\bar{m}\cdot\log(\NumBins) + \NumBins\big).
\]
\par
\Tb{Case 2:} The inner boundaries $a_{1},\ldots,a_{\NumBins-1}$ are fixed like
when using $a^{\ast}$: Let $x=(x_{1},\ldots,x_{\bar{m}})$. For every $i$ we
can compute the unique bin $[a_{k},a_{k+1})$ ($k\geqq1$) containing $x_{i}$ by
first computing the constant bin width $w=\frac{\max(x)-\min(x)}{\NumBins-2}$,
then $k=1+\lfloor\frac{x_{i}-\min(x)}{w}\rfloor$.\footnote{Note that the
  right-most inner interval must be
  $[a_{\NumBins-2},a_{\NumBins-1}+\varepsilon)$ instead of
  $[a_{\NumBins-2},a_{\NumBins-1})$ for
  $a_{\NumBins-1}=\max(x_{1},\ldots,x_{\bar{m}})$ for formal reasons, see
  subsection \ref{sssec:HistogramFunctions}.}
Obviously, $k\in\{1,\ldots,\NumBins-1\}$ and $x_{i}\in[a_{k},a_{k+1})$.  The
left-most interval $[a_{0},a_{1})=[-\infty,a_{1})$ stays empty. We need to
compute $\min(x)$ and $\max(x)$ only once and have $\McO(1)$ arithmetic
operations plus incrementing one of $\NumBins$ counters for every $x_{i}$
which sums up to
\[  \McO(\bar{m}+\NumBins)
\]
operations altogether.
\\
Note that in our applications $\bar{m}$ is $|T^{+}|$ or $|T^{-}|$ which are
both $\leqq m$ (= row number of $X$).
\subsection{Computing
  $\Mr{FindCics}(\Mr{Scale}(X),T^{+},T^{-},b^{+},b^{-},\NumBins)$ (Algorithm
  \ref{alg:FindCics})}
\label{ssec:ComplexityOfFindCics}
For every column index $j=1,\ldots,n$: Computing $\min(x^{+})$ and
$\max(x^{+})$ takes $\McO(|T^{+}|)$ operations, thus computing $a^{\ast}$ is
in $\McO(|T^{+}|+\NumBins)$.  Computing the histograms $h^{+}$ and $h^{-}$
takes $\McO(|T^{+}|) + \McO(|T^{-}|) + \McO(\NumBins)$ operations according to
Case 2 of above. Computing $h_{\max}^{+}$ and $k^{+}$ takes $\NumBins$
comparisons at most.  The output consists of at most $3\cdot n$ numbers.
\par
So computing $\Mr{FindCics}(\Mr{Scale}(X),T^{+},T^{-},b^{+},b^{-},\NumBins)$
from $\Mr{Scale}(X)$ takes at most
\[ \McO\big(n\cdot (|T^{+}|+|T^{-}|+\NumBins)\big)
\]
operations.
\par
\Tb{Remark.} This is in $\McO\big(n\cdot(m+\NumBins)\big)$ since
$|T^{+}|+|T^{-}|\leqq m$.
\par
\Tb{Theorem.} Computing
$\Ms{C}=\Mr{FindCics}(\Mr{Scale}(X),T^{+},T^{-},b^{+},b^{-},\NumBins)$ from
$X\in\MbQ^{m\times n}$ takes at most
\[  \McO\big(n\cdot(m+\NumBins)\big)
\]
operations.
\par
\Tb{Proof.} This follows directly from the complexity of computing
$\Mr{Scale}(X)$, then computing
$\Ms{C}=\Mr{FindCics}(\Mr{Scale}(X),T^{+},T^{-},b^{+},b^{-},\NumBins)$ from
$\Mr{Scale}(X)$ and from the above Remark, qed.
\subsection{Computing $\Ms{S}_{\Ms{C}}(I_{\neg T})$ (Algorithm \ref{alg:ComputingIndicatorValues})}
Given
$\Ms{C}=\Mr{FindCics}(\Mr{Scale}(X),T^{+},T^{-},b^{+},b^{-},\NumBins)$, for
each one of $I_{\neg T}$ non-training objects we must perform at most
$|\Ms{C}|$ interval containment decisions which sums up to
\[  |\Ms{C}|\cdot\McO(|I_{\neg T}|)
\]
operations.
\subsection{Computing predictions $F(i,c)$ (Algorithm
    \ref{alg:ComputingPredictions})}
Given $\Ms{S}_{\Ms{C}}(I_{\neg T})$ and $c$, this sums up to 
\[  \McO(|I_{\neg T}|)
\]
operations for all $i\in I_{\neg T}$.
\subsection{Na\"ive cutoff selection}
Given $I^{+}_{\neg T}$, $I^{-}_{\neg T}$ and $\Ms{S}_{\Ms{C}}(I_{\neg T})$
from Algorithm \ref{alg:ComputingIndicatorValues}, Computing $Av^{0},Av^{1}$
takes
\[  \McO(|I^{+}_{\neg T}|+|I^{-}_{\neg T}|)\subseteq\McO(|I_{\neg T})
\]
operations.
\subsection{Optimizing cutoff $c$ (Algorithm \ref{alg:OptimizingCutoff})}
Let $Q:\MbB^{r}\times\MbB^{r}\longrightarrow\MbR$ be a function measuring some
type of similarity of two 0-1 vectors and let be $E(r)$ the maximum number of
operations it takes to evaluate $Q(v,w)$ for any $v,w\in\MbB^{r}$.  Given
$\Ms{C}=\Mr{FindCics}(\Mr{Scale}(X),\allowbreak
T^{+},T^{-},b^{+},b^{-},\NumBins)$ from Algorithm \ref{alg:FindCics},
Algorithm \ref{alg:OptimizingCutoff} repeats
$\left(\Ms{S}_{\Ms{C}}^{\max}-\Ms{S}_{\Ms{C}}^{\min}+1\right)$ times Algorithm
\ref{alg:ComputingPredictions} for obtaining a prediction vector $\Ms{F}(c)$
using $\McO(|I_{\neg T}|)$ operations each, and evaluates $Q(\cdot,\cdot)$ on
two 0-1 vectors of length $r=|I_{\neg T}|$ using at most $E(|I_{\neg T}|)$
operations each.  From $\Ms{S}_{\Ms{C}}^{\min}\geqq0$ and
$\Ms{S}_{\Ms{C}}^{\max}\leqq|\Ms{C}|$ follows
$\Ms{S}_{\Ms{C}}^{\max}-\Ms{S}_{\Ms{C}}^{\min}+1\leqq |\Ms{C}|+1$.  Computing
$Q_{opt}$ and $C_{opt}$ both take
$\McO(\Ms{S}_{\Ms{C}}^{\max}-\Ms{S}_{\Ms{C}}^{\min}+1)$ as $c$ takes just this
many values in the for-loop.
\par
So the total number of operations Algorithm \ref{alg:OptimizingCutoff} takes
is limited by
\begin{eqnarray*}
  (|\Ms{C}|+1)\cdot\McO\left(|I_{\neg T}| + E(|I_{\neg T}|) \right)  \\
  \subseteq
     \McO\left( |\Ms{C}|\cdot\big(|I_{\neg T}| + E(|I_{\neg T}|)\big) \right) 
\end{eqnarray*}
\subsection{Overall complexity }
Adding the upper bounds for $\Mr{Scale}(X)$, for computing
$\Mr{FindCics}\big(\Mr{Scale}(X),\ldots\big)$ and for computing
$\Ms{S}_{\Ms{C}}(I_{\neg T})$ gives an operation count of
\[
\begin{array}{lll}
  & \McO\big(n\cdot m + n\cdot(m+\NumBins) + |\Ms{C}|\cdot|I_{\neg T}|\big) \\
  \subseteq& \McO\big(n\cdot(m + \NumBins + |I_{\neg T}|)\big)\qquad
              & \text{(since $|\Ms{C}|\leqq n$)} \\
  \subseteq& \McO\big(n\cdot(m + \NumBins)\big)\qquad
              & \text{(since $|I_{\neg T}|\leqq m$)}
\end{array}
\]
If we add steps for cutoff selection or optimization and use
$|I_{\neg T}|\leqq m$ twice we get:
\begin{eqnarray*}
      & \McO\big(n\cdot(m + \NumBins)\big)
          & \text{with na\"ive cutoff selection} \\
      & \McO\Big(n\cdot\big(m + \NumBins + E(|I_{\neg T}|)\big)\Big)
          & \text{when optimizing cutoff $c$}
\end{eqnarray*}
Using $|I_{\neg T}|\leqq m$ once again, we get:
\par
\Tb{Theorem.} Given $X\in\MbQ^{m\times n}$, $v\in\MbB^{m}$, training sets
$T^{+},T^{-}\subseteq\{1,\ldots,m\}$,
$b^{+},b^{-}\in\MbQ,\NumBins\in\MbN_{\geqq3}$, the overall number of
operations it takes to
\begin{itemize}
\item
  compute $\Mr{Scale}(X)$
\item
  compute $\Ms{C}=\Mr{FindCics}\big(\Mr{Scale}(X),\allowbreak
  T^{+},T^{-},b^{+},b^{-},\NumBins\big)$
\item
  compute $\Ms{S}_{\Ms{C}}(I_{\neg T})$\qquad
      \big(where
        $I_{\neg T} = \{1,\ldots,m\}\setminus (T^{+}\cup T^{-})$\big)
\item
  select cutoff $c$
\item
  compute predictions $F(i,c)$ from this by Algorithm
  \ref{alg:ComputingPredictions}
\end{itemize}
is limited by
\begin{eqnarray*}
      & \McO\big(n\cdot(m + \NumBins)\big)
          & \text{with na\"ive cutoff selection} \\
      & \McO\Big(n\cdot\big(m + \NumBins + E(m)\big)\Big)
          & \text{when optimizing cutoff $c$}
\end{eqnarray*}
where $E(m)$ is the number of operations it takes to evaluate the chosen
similarity measure $Q:\MbB^{m}\times\MbB^{m}\longrightarrow\MbR$.
\par
\Tb{Note.} The direct input of Algorithms \ref{alg:ComputingIndicatorValues}
and \ref{alg:OptimizingCutoff} is $\Ms{C}$, not $X$ or $\Mr{Scale}(X)$, so
their complexity is actually linear in the number of Cics $|\Ms{C}|$.  In the
worst case $|\Ms{C}|$ equals $n$ but in practical settings $|\Ms{C}|$ is
frequently much smaller than $n$, for example $20$ versus $150$.  So
Algorithms \ref{alg:ComputingIndicatorValues} and \ref{alg:OptimizingCutoff}
become the faster the smaller the number of Cics $|\Ms{C}|$ is.  Indeed, the
implementation on a real computer shows that $|\Ms{C}|$ is an important
determinant for time consumption.

\section{Uniting Two Classification Algorithms}
\label{sec:Uniting}
There are several ways to combine multiple raters with the goal of obtaining a
better rater.  Some well-known ways are making majority decisions, or making a
decision according to the number of the ``1'' (or ``0'') of the single raters
exceeding a certain threshold or not.  What we are doing in this section
instead is combining the baselying principles of two classification
algorithms---Algorithm \ref{alg:Classification} and the one of
\cite{Olschewski:2021}---into one algorithm.
\par
In this section, we will use the same notations and designations as in section
\ref{sec:TheAlgorithms}
\par
\Tb{Definition.} We will say that a Cic $j$ is \Tb{active} for object $i$
represented by $x_{i,1},\ldots,x_{i,n}$ if in Algorithm
\ref{alg:ComputingIndicatorValues},
$x_{i,j}\in\left[a^{\ast}_{k^{+}},a^{\ast}_{k^{+}+1}\right)$.  Here,
$i\in\{1,\ldots,m\}$, $j\in\{1,\ldots,n\}$.
\par
In $\Ms{S}_{\Ms{C}}(i)$ computed by Algorithm
\ref{alg:ComputingIndicatorValues}, the number of active Cics for object $i$
is used as an indicator for positivity of object $i$.  Actually, there is no
reason for just counting and not trying to go further.  In order to be more
specific, one could replace $\Ms{S}_{\Ms{C}}(i)$ by the bit pattern induced by
the active Cics for object $i$.
\par
\Tb{Definition.} Let
\begin{align*}
  \Ms{A}_{\Ms{C}}(i):\{1,\ldots,m\} & \to\MbB^{|\Ms{C}|} \\
  i & \mapsto a_{i}
\end{align*}
where
\[ a_{i}(j) =
  \begin{cases}
    1 & \text{if Cic $j$ is active for object $i$} \\
    0 & \text{else}
  \end{cases}
\]
Now let be $U^{+}\subseteq I^{+}$ some set of positive objects we will use as
a third set of training indices along with $T^{+},T^{-}$. We need to reduce
$I_{\neg T}$ defined in subsection \ref{sssec:HistogramFunctions} to
\[
    I_{\neg T,\neg U}=\{1,\ldots,m\}\setminus\left(T^{+}\cup T^{-}\cup
    U^{+}\right)
\]
in order to skip the objects of three training sets in
the classification loop.
\par
Let be $\langle a,b\rangle=\sum_{k=1}^{r} a_{k}\cdot b_{k}$ the inner product
of two vectors of size $r$, and $H(a)=\sum_{k=1}^{r}a_{k}$ the Hamming weight
of $a\in\MbB^{r}$.
\par
\Tb{Definition.}  Let be $i\in\{1,\ldots,m\}$ and $U^{+}$ as
above.
\begin{eqnarray*}
  q_{\max}(i,U^{+}) &:=& \max_{j\in U^{+}}\frac{\langle \Ms{A}_{\Ms{C}}(i),
            \Ms{A}_{\Ms{C}}(j)\rangle}{H\left(\Ms{A}_{\Ms{C}}(i)\right)} \\
  q_{\min}(i,U^{+}) &:=& \min_{j\in U^{+}}\frac{\langle \Ms{A}_{\Ms{C}}(i),
            \Ms{A}_{\Ms{C}}(j)\rangle}{H\left(\Ms{A}_{\Ms{C}}(i)\right)}
\end{eqnarray*}
\par
\Tb{Remark.} $q_{\max}(i,U^{+})$ is in strict analogy to what is denoted by
$s(x,T)$ in subsection ``Dense Formulation'' of section ``The Algorithms'' in
\cite{Olschewski:2021}, the difference being that in \cite{Olschewski:2021} we
operate on bit patterns $\left(P_{1}(x^{(i)}),\ldots,P_{n}(x^{(i)})\right)$
induced by component-wise thresholding $x^{(i)}=\text{$i$-th line of the
  scaled input matrix}$, whereas by computing $q_{\max},q_{\min}$ here we
operate on the Cic activity pattern of line $i$ of the (scaled) input matrix.
\par
\Tb{Definition.} As in subsection \ref{sssec:ComputingIndicatorValues}, let
$\Ms{C}=\Mr{FindCics}(\Mr{Scale}(X),T^{+},T^{-},b^{+},b^{-},\NumBins)$ be the
output of Algorithm \ref{alg:FindCics}.  Define two alternative indicators by:
\begin{align*}
  \Ms{Q}^{\max}_{\Ms{C},U^{+}} & : I_{\neg T,\neg U}
      \to [0,1]\subseteq\MbQ \\
  i &\mapsto q_{\max}(i,U^{+}) \\
  \Ms{Q}^{\min}_{\Ms{C},U^{+}} & : I_{\neg T,\neg U}
      \to [0,1]\subseteq\MbQ \\
  i &\mapsto q_{\min}(i,U^{+}) \\
\end{align*}
\par
The following algorithm is formulated in a way that it can be used as a
drop-in replacement for Algorithm \ref{alg:ComputingIndicatorValues} in
subsection \ref{sssec:ComputingIndicatorValues}.
\\~\\
\begin{algorithm}[H]
  \label{alg:ComputingQ}
  \KwInput{$\Ms{C}=\Mr{FindCics}(\Mr{Scale}(X),T^{+},T^{-},U^{+},b^{+},b^{-},
    \NumBins)$}
  \KwOutput{$\Ms{Q}^{\max}_{\Ms{C},U^{+}}\left(I_{\neg T,\neg U}\right),
    \Ms{Q}^{\min}_{\Ms{C},U^{+}}\left(I_{\neg T,\neg
        U}\right)\in\MbQ^{|I_{\neg T, \neg U}|}$}
  \caption{Computing alternative indicators
    $\Ms{Q}^{\max}_{\Ms{C},U^{+}}(i),\Ms{Q}^{\min}_{\Ms{C},U^{+}}(i)$}
  \For{$i\in I_{\neg T,\neg U}$}
  {
    compute $\Ms{Q}^{\max}_{\Ms{C},U^{+}}(i)$ \\
    compute $\Ms{Q}^{\min}_{\Ms{C},U^{+}}(i)$
  }
\end{algorithm}
~\\
We can now formulate a variant of the classification Algorithm
\ref{alg:Classification} of subsection \ref{sssec:ClassificationAlgorithm}
which uses
$\Ms{Q}^{\max}_{\Ms{C},U^{+}}\left(I_{\neg T,\neg U}\right),
\Ms{Q}^{\min}_{\Ms{C},U^{+}}\left(I_{\neg T,\neg U}\right)$ as indicators
instead of $\Ms{S}_{\Ms{C}}\left(I_{\neg T}\right)$.
\\~\\
\begin{algorithm}[H]
  \label{alg:UnitingClassification}
  \KwInput{$X\in\MbQ^{m\times n};b^{+},b^{-}\in\MbQ;
    \NumBins\in\MbN_{\geqq3};t^{+},t^{-},u^{+}\in\MbN_{\geqq1}$}
  \KwOutput{$\Ms{F}(C_{opt})\in\MbB^{|I_{\neg T,\neg U}|}$}
  \caption{Classification algorithm}
  $\Mr{Scale(X)}$  by
    Algorithm \ref{alg:Scale}\\
    select random subsets $T^{+}\subseteq I^{+}$ of size $t^{+}$,
    $T^{-}\subseteq I^{-}$ of size $t^{-}$,
    $U^{+}\subseteq I^{+}$ of size $u^{+}$ \\
  $\Ms{C}:=\Mr{FindCics}(\Mr{Scale}(X),T^{+},T^{-},b^{+},b^{-},\NumBins)$ by
      Algorithm \ref{alg:FindCics} \\
  compute $\Ms{Q}^{\max}_{\Ms{C},U^{+}}\left(I_{\neg T,\neg U}\right)$,
      $\Ms{Q}^{\min}_{\Ms{C},U^{+}}\left(I_{\neg T,\neg U}\right)$ by
      Algorithm \ref{alg:ComputingQ} \\
  compute the optimum cutoff $C_{opt}$ and $\Ms{F}(C_{opt})$ by an adaption of
      Algorithm \ref{alg:OptimizingCutoff} and Algorithm
      \ref{alg:ComputingPredictions}
\end{algorithm}
~\\
By ``an adaption of Algorithm \ref{alg:OptimizingCutoff} and Algorithm
\ref{alg:ComputingPredictions}'' we mean the following two modifications:
\begin{itemize}
\item[$-$] In Algorithm \ref{alg:OptimizingCutoff}, replacing the looping
  $c=\Ms{S}^{\min}_{\Ms{C}},
  \Ms{S}^{\min}_{\Ms{C}}+1,\ldots,\Ms{S}^{\max}_{\Ms{C}}$ by looping over some
  grid $\mathcal{C}\subset[0,1]\subset\MbQ$ or by the method of subsection
  ``Refined Method For Cutoff Selection'' in \cite{Olschewski:2021}.
\item[$-$] In Algorithm \ref{alg:ComputingPredictions} (called by Algorithm
  \ref{alg:OptimizingCutoff}), replacing the computation of
  $\Ms{S}_{\Ms{C}}(i)$ by $\Ms{Q}^{\max}_{\Ms{C},U^{+}}(i)$,
  $\Ms{Q}^{\min}_{\Ms{C},U^{+}}(i)$ as defined above.
\end{itemize}
The potential of Algorithm \ref{alg:UnitingClassification} has yet to be
explored and will be a subject of future research.

\section{Conclusion}
\label{sec:Conclusion}
We have presented a histogram-based algorithm for object classification in
subsection \ref{sssec:ClassificationAlgorithm} together with some worst-case
complexity analysis in section \ref{sec:AlgorithmicComplexity}, and rendered
the detection and prediction of frontend and backend defects in semiconductor
chips based on wafer measurement data as main application by discussing
classification results of data from real-world wafer fabrication in section
\ref{sec:Results}.  Furthermore, we laid out a method of searching for
candidate indicator columns in subsection \ref{ssec:DimensionalReduction}
which enables the feature count to be reduced dramatically in some
applications.  We presented an algorithm using $n_{\Mr{diff}}$ as a feature
relevance indicator in section \ref{ssec:AutomatizingTheFindingofCics}.
\par
Finally, we presented an algorithm that unites some baselying principle of the
classification algorithm of \cite{Olschewski:2021} with the principles of the
classification Algorithm \ref{alg:Classification} as detailed in this paper.
\par
In subsection \ref{ssec:PredictingBackendDefectStatusofChips} of this paper,
we have started tackling the problem of predicting the backend overall defect
states of semiconductor chips which has been mentioned in the last paragraph
of \cite{Olschewski:2021} as one topic of ongoing research.  The results which
we report on in section \ref{sec:Results} confirm that predicting the backend
overall defect state for the type of data our classifications are performed on
using only frontend measurements seems to be harder of a machine learning
problem than predicting the frontend overall defect state.

\listoffigures
\bibliography{match-mode}{}
\bibliographystyle{plain}
\end{document}